\documentclass{article}

\PassOptionsToPackage{numbers, compress}{natbib}

\usepackage[final]{neurips_2023}

\usepackage[utf8]{inputenc} %
\usepackage[T1]{fontenc}    %
\usepackage[pagebackref=true,unicode=true,colorlinks=true,citecolor=blue,linkcolor=red]{hyperref}
\usepackage{url}            %
\usepackage{booktabs}       %
\usepackage{colortbl}
\usepackage{amsfonts}       %
\usepackage{nicefrac}       %
\usepackage{microtype}      %
\usepackage{pifont}%
\usepackage[dvipsnames]{xcolor}
\usepackage{amsmath}
\usepackage{graphicx}
\usepackage{pgfplots}
\usepackage{comment}
\usepackage{amsmath,amssymb} %
\usepackage{color}
\usepackage{soul}
\usepackage{svg}
\usepackage{wrapfig}
\usepackage{subfig}
\usetikzlibrary{patterns}
\setcitestyle{square}

\usepackage{multirow}
\usepackage{appendix}

\usepgfplotslibrary{groupplots}
\pgfplotsset{compat=1.13}

\def\ours{\texttt{POP-3D}} %

\newcommand{\PP}[1]{\textcolor{black}{#1}}

\newcommand{\AV}[1]{\textcolor{black}{#1}}

\newcommand{\crd}[1]{\textcolor{black}{#1}}

\definecolor{free}{RGB}{110, 140, 255}

\newcommand{\voxft}{\mathbf{V}}
\newcommand{\voxext}{f_\mathrm{3D}}
\newcommand{\imext}{f_\mathrm{I}}
\newcommand{\occhead}{g}
\newcommand{\fthead}{h}
\newcommand{\task}{3D occupancy prediction}
\newcommand{\Occ}{\mathbf{O}}
\newcommand{\Targ}{\mathbf{T}}

\DeclareMathOperator*{\argmax}{arg\,max}

\title{\ours: Open-Vocabulary 3D Occupancy Prediction from Images}

\author{%
  Antonin~Vobecky$^{1,2,3}$
\And
Oriane Siméoni$^1$
\And
David Hurych$^1$
\And 
Spyros Gidaris$^1$
\And 
Andrei Bursuc$^1$
\And
Patrick Pérez$^1$
\And
 Josef~Sivic$^2$
\AND
\normalfont $^1$\,valeo.ai, Paris, France \quad $^2$\, \AV{CIIRC CTU in Prague} \quad $^3$\, \AV{FEE CTU in Prague}
}

\newcommand\fullname[1]{%
  \begingroup
  \renewcommand\thefootnote{}\footnote{#1}%
  \addtocounter{footnote}{-1}%
  \endgroup
}

\begin{document}

\maketitle

\begin{abstract}
  We describe an approach to predict open-vocabulary 3D semantic voxel occupancy map from input 2D images with the objective of enabling 3D grounding, segmentation and retrieval of free-form language queries. This is a challenging problem because of the 2D-3D ambiguity and the open-vocabulary nature of the target tasks, where obtaining annotated training data in 3D is difficult. The contributions of this work are three-fold. 
First, we design a new model architecture for open-vocabulary 3D semantic occupancy prediction.  The architecture consists of a 2D-3D encoder together with occupancy prediction and 3D-language heads. The output is a dense voxel map of 3D grounded language embeddings enabling 
a range of open-vocabulary tasks. 
Second, we develop a \emph{tri-modal} self-supervised learning algorithm that leverages three modalities: (i) images, (ii) language and (iii) LiDAR point clouds, and enables 
training the proposed architecture using a strong pre-trained vision-language model without the need for any 3D manual language annotations. 
Finally, we demonstrate quantitatively the strengths of the proposed model on several open-vocabulary tasks:
Zero-shot 3D semantic segmentation using existing datasets; 3D grounding and retrieval of free-form language queries, using a small dataset that we propose as an extension of nuScenes. You can find the project page here \href{https://vobecant.github.io/POP3D}{\texttt{https://vobecant.github.io/POP3D}}.
\setcounter{footnote}{1}
\fullname{$^2$Czech Institute of Informatics, Robotics and Cybernetics at the Czech Technical University in Prague}

\end{abstract}

\section{Introduction}

The detailed analysis of 3D environments --both geometrically and semantically-- is a fundamental perception brick in many applications, from augmented reality to autonomous robots and vehicles. It is usually conducted with cameras and/or laser scanners (LiDAR). In its most complete version, called \textit{semantic 3D occupancy} prediction, this analysis amounts to labeling each voxel of the perceived volume as occupied by a particular class of object or empty. This is extremely challenging since both cameras and LiDAR only capture information about visible surfaces, which may be projected from 3D into 2D without the loss of information, but not for every point in the 3D space. This one extra dimension makes prediction arduous and hugely complicates the manual annotation task.

Recent works, e.g., \cite{huang2023tri}, propose to leverage manually annotated LiDAR data to produce a partial annotation of the 3D occupancy space. However, relying on manual semantic annotation of point clouds remains challenging to scale, even if sparse, and limits the learned representation to encode solely a closed vocabulary, i.e., a limited predefined set of classes. In this work, we tackle these challenges and propose an open-vocabulary approach to 3D semantic occupancy prediction that relies only on unlabeled image-LiDAR data for training. In addition, our model uses only camera inputs at run time, bypassing the need for expensive dense LiDAR sensors altogether, in contrast with most 3D semantic perception systems (whether at point or voxel level). 

\begin{figure}[t!]
    \centering
    \includegraphics[width=1.\linewidth]{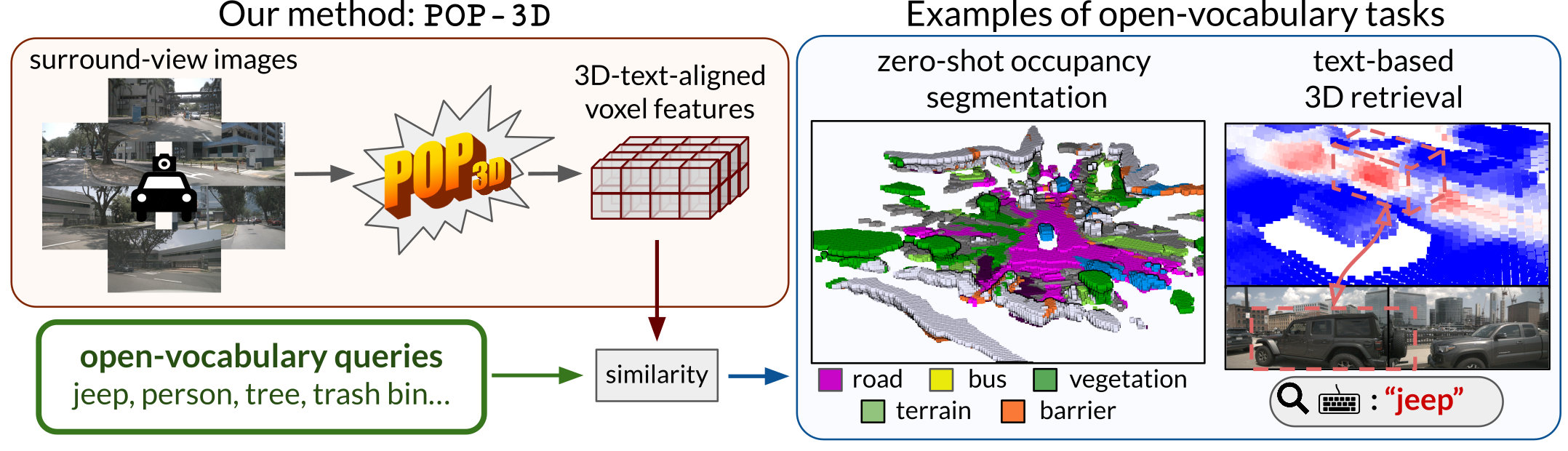}
    \caption{\textbf{Overview of the proposed method.} Provided only with surround-view images as input, our model called~\ours~produces a voxel grid of 3D text-aligned features that support open-vocabulary downstream tasks such as zero-shot occupancy segmentation or text-based grounding and retrieval. 
    } 
    \label{fig:teaser}
\end{figure}

To this end, we 
harness the progress made recently in supervised 3D occupancy prediction \cite{huang2023tri} and in language-image alignment \cite{zhou2022maskclip} within a two-head image-only model that can be trained with aligned image-LiDAR raw data. We first train a class-agnostic occupancy prediction head to leverage the sparse 3D occupancy information that LiDAR scans provide for free. Using this same LiDAR information and pre-trained language-aligned visual features at the corresponding locations in images, we jointly train a second head that predicts the same type of features at the 3D voxel level. At run time, these features can be probed from text prompts to get open-vocabulary semantic segmentation of voxels predicted as occupied (\autoref{fig:teaser}).  
To assess the effectiveness of our method for semantic 3D occupancy prediction, we introduce a novel evaluation protocol 
specifically tailored to this task. By evaluating this protocol on autonomous driving data, our method achieves a strong performance 
relative to the fully-supervised approach.

In a nutshell, we attack the complex problem of 3D semantic occupancy prediction with the lightest possible requirements: no manual annotation of the training data, no predefined semantic vocabulary, and no recourse to LiDAR readings at run time. As a result, the proposed image-only 3D semantic occupancy model 
named \ours~(for o\textbf{P}en-vocabulary \textbf{O}ccupancy \textbf{P}rediction in \textbf{3D}) provides training data scalability and operational versatility while opening up new understanding capabilities for autonomous systems through language-driven scene perception.

\section{Related work}

\textbf{Semantic 3D occupancy prediction.~}
Automatic understanding of the 3D geometry and semantics of a scene has been traditionally enabled through high-precision LiDAR sensors and corresponding architectures. 
3D semantic segmentation, i.e., point-level classification of a point cloud, can be addressed with different types of transformations of the point cloud: point-based, directly operating on the three-dimensional points \cite{qi2017pointnet,qi2017pointnet++,thomas2019kpconv}, and projection-based, operating on a different representation, \emph{e.g.}, two-dimensional images \cite{wu2019squeezesegv2,lawin2017deep,Boulch2017UnstructuredPC} or three-dimensional voxel representations~\cite{yan20222dpass,zhu2021cylindrical,tang2020searching,minkowskichoy20194d}. 
However, they produce predictions as sparse as the LiDAR point cloud, offering an incomplete understanding of the whole scene. Semantic scene completion~\cite{roldao20223d} aims for dense inference of 3D geometry and semantics of objects and surfaces within a given extent, typically leveraging rich geometry information at the input extracted from depth~\cite{chen2020bsp,li2020anisotropic}, occupancy grids~\cite{wu2020scfusion, roldao2020lmscnet}, point clouds~\cite{rist2021semantic}, or a mix of modalities, e.g., RGBD~\cite{cai2021semantic, cherabier2018learning}. 
In this line, MonoScene~\cite{cao2022monoscene} is the first camera-based method to produce dense semantic occupancy predictions from a single image by projecting image features into 3D voxels by optical ray intersection. Recent progress in multi-camera Bird's-Eye-View (BEV) projection~\cite{philion2020lift,huang2021bevdet,zhou2022cross,li2022bevformer, bartoccioni2022lara,li2022bevdepth} enables the recent TPVFormer~\cite{huang2023tri} to generate surrounding 3D occupancy predictions by effectively exploiting tri-perspective view representations~\cite{chan2022efficient} augmenting the standard BEV with two additional perpendicular planes to recover the full 3D. All prior methods are trained in a supervised manner, requiring rich voxel-level semantic information, which is costly to curate and annotate.
While we build on~\cite{huang2023tri}, we forego manual label supervision and, instead, develop a model able to produce semantic 3D occupancy predictions using supervision from LiDAR and from an image-language model, allowing our model to acquire open-vocabulary skills in the voxel space.

\textbf{Multi-modal representation learning.~}
Distilling signals and knowledge from one modality into another is an effective strategy to learn representations~\cite{alayrac2020self,alwassel_2020_xdc} or to learn to solve tasks using only few~\cite{chen2021localizing, owens2018audio, afouras2021self} or no human labels~\cite{tian2021unsupervised,vobecky2022drive}. The interplay between images, language, and sounds is often used for self-supervised representation learning over large repositories of unlabeled data fetched from the internet~\cite{alayrac2020self,alwassel_2020_xdc,arandjelovic2017look,miech2020end,owens2018audio}. Images can be paired with different modalities towards solving complex 2D tasks, e.g., semantic segmentation~\cite{vobecky2022drive}, detection of road objects~\cite{tian2021unsupervised} or sound-emitting objects ~\cite{chen2021localizing, owens2018audio, afouras2021self}. Image-language aligned models project images and text into a shared representation space~\cite{gordo2017,sariyildiz2020,unicoder2020,li2021,desai2021, radford2021learning, Jia2021}. Contrastive image-language learning on many millions of image-text pairs~\cite{radford2021learning,Jia2021} leads to high-quality representations with impressive zero-shot skills from one modality to the other. 
We use CLIP~\cite{radford2021learning} for its appealing open-vocabulary property that enables the querying of visual content with natural language toward recognizing objects of interest without manual labels. 
\ours~uses LiDAR supervision for precise occupancy prediction and learns to produce in the 3D space CLIP-like features easily paired with language.

{\textbf{Open-vocabulary semantic segmentation.~}
Zero-shot semantic segmentation aims to segment object classes not seen during training~\cite{xian2019semantic,bucher2019zero,hu2020uncertainty}. The advent of CLIP~\cite{radford2021learning}, which is trained on abundant web data, has inspired a new wave of methods, dubbed \emph{open-vocabulary}, for recognizing random objects via natural language queries. CLIP features can be projected into 3D meshes~\cite{conceptfusion} and NeRFs~\cite{lerf2023} to enable language queries. Originally producing image-level embeddings, CLIP can be extended to pixel-level predictions for open-vocabulary semantic segmentation by exploiting different forms of supervision from segmentation datasets, e.g., pixel-level labels~\cite{li2022languagedriven} or class agnostic masks ~\cite{ghiasi2022scaling, liang2022open, zhong2022regionclip} coupled with region-word grounding~\cite{grounding}, however with potential forgetting of originally learned concepts~\cite{conceptfusion}. MaskCLIP+~\cite{zhou2022maskclip} adjusts the attentive-pooling layer of CLIP to generate pixel-level CLIP features that are further distilled into an encoder-decoder semantic segmentation network. MaskCLIP+~\cite{zhou2022maskclip} preserves the open-vocabulary properties of CLIP, and we exploit it here to distill its knowledge into \ours. We generate target 3D CLIP features by mapping MaskCLIP+ pixel-level features to LiDAR points observed in images.
By being trained to match these distillation targets, \ours~manages to learn 3D features with open-vocabulary perception abilities, in contrast to prior work on \task ~that is limited to recognizing a closed-set of visual concepts.

\section{Open-vocabulary 3D occupancy prediction}

Our goal is to predict 3D voxel representations
of the environment, given a set of 2D input RGB images, that is amenable to open-vocabulary tasks such as zero-shot semantic segmentation or concept search driven by natural language queries.
This is a challenging problem as we need to address the following two questions. First, what is the right architecture to handle the 2D-to-3D ambiguity and the open-vocabulary nature of the task? Second, how to formulate the learning problem without requiring manual annotation of
large amounts of 3D voxel data, which are extremely hard to produce.%

To address these questions, we propose the following two innovations. First, we design an architecture for open-vocabulary 3D occupancy prediction (\autoref{fig:proposed_approach}(a) and \autoref{sec:architecture}) that handles the 2D-to-3D prediction and open-vocabulary tasks with two specialized heads. Second, we formulate its training as a \textit{tri-modal self-supervised learning} problem (\autoref{fig:proposed_approach}(b) and \autoref{sec:trimodal_learning}) that leverages aligned (i) 2D images with (ii) 3D point clouds equipped with (iii) pre-trained language-image features as the three input modalities (i.e. camera, LiDAR and language) without the need for any explicit manual annotations. The details of these contributions are given next.

\begin{figure}[t]
    \centering
    \includegraphics[width=1.0\linewidth]{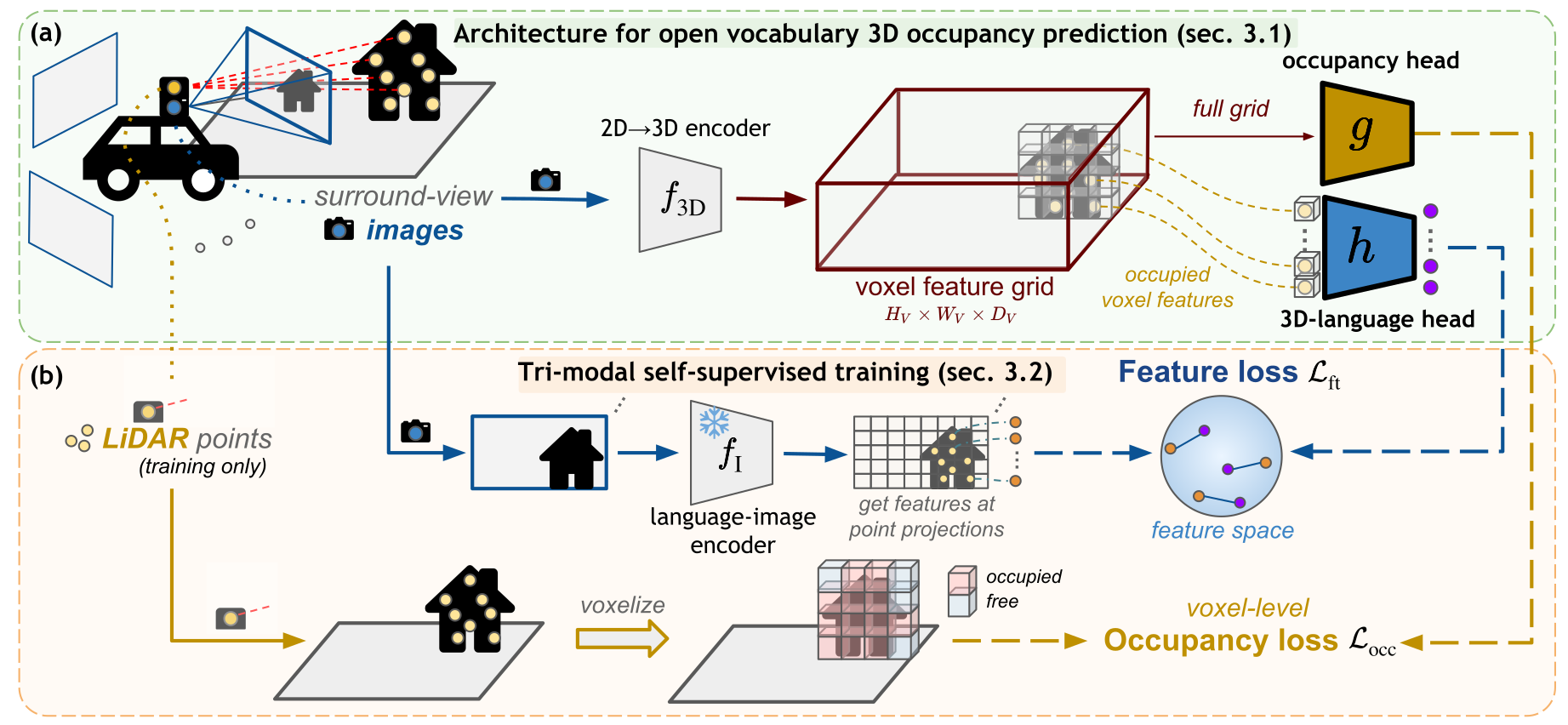}
    \caption{
    \textbf{Proposed approach.} 
    In \textbf{(a)}, we show the architecture of the proposed method. Having only surround-view images on the input, the model first extracts a dense voxel feature grid that is then fed to two parallel heads: occupancy head~$\occhead$ producing voxel-level occupancy predictions, and to 3D-language feature head~$\fthead$ which outputs features aligned with text representations.
    In \textbf{b)}, we show how we train our approach, namely the occupancy loss~$\mathcal{L}_\text{occ}$ used to train class-agnostic occupancy predictions, and the feature loss~$\mathcal{L}_\text{ft}$ that enforces the 3D-language head~$\fthead$ to output features aligned with text representations.
    }
    \label{fig:proposed_approach}
\end{figure}

\subsection{Architecture for open-vocabulary 3D occupancy prediction}
\label{sec:architecture}

We are given a set of surround-view images captured from one camera location, and our goal is to output a 3D occupancy voxel map and support language-driven tasks. 
To reach the goals, we propose an architecture composed of three modules (\autoref{fig:proposed_approach}(a)). First, a \textit{2D-3D encoder} predicts a voxel feature grid from the input images. Second, the \textit{occupancy head} decodes this entire voxel grid into an occupancy map, predicting which voxels are free and which are occupied. Finally, the \textit{3D-language head} is applied on each occupied voxel to output a powerful language embedding vector enabling a range of 3D open-vocabulary tasks. 
The three modules are described next.

\textbf{2D-to-3D encoder $\voxext$.~} 
The objective of the 2D-to-3D encoder is to predict a dense feature voxel grid given one or more images captured at one location as input. 
The output voxel grid representation encodes 3D visual information captured by the cameras. 
In detail, given surround-view camera RGB images 
$\mathbf{I}$ and camera calibration parameters, 
the encoder~$\voxext$ produces a feature voxel grid %
\begin{equation}
\voxft=\voxext\left(\mathbf{I}\right)\in\mathbb{R}^{H_\text{V}\times W_\text{V}\times D_\text{V}\times C_\text{V}},
\end{equation}
where $H_\text{V},W_\text{V},$ and $D_\text{V}$ are the spatial dimensions of the voxel grid, and 
$C_\text{V}$ is the feature dimension of each voxel.
This feature voxel grid is then passed to two distinct prediction heads designed to perform \textit{class-agnostic occupancy prediction} and \textit{text-aligned feature prediction} tasks, respectively. The two heads are described next.

\textbf{Occupancy head $g$.~} 
Given the feature voxel grid $\voxft$, the occupancy prediction head $\occhead$ aims at classifying every voxel as `empty' or `occupied.' 
Following~\cite{huang2023tri}, this head is implemented as a non-linear network composed of $N_\text{occ}$ hidden blocks with configuration \texttt{Linear-Softplus-Linear}, each with $C_\text{occ}^\text{hidden}$ hidden features, and a final linear classifier outputting two logits, one per class. It outputs the tensor
\begin{equation}
\Occ_\text{occ} = \occhead \left( \voxft \right) \in \mathbb{R} ^ {H_\text{V}\times W_\text{V}\times D_\text{V}\times 2}, 
\end{equation}
containing the occupancy prediction for each voxel. %

\textbf{3D language head $h$.~} In parallel, the voxel grid $\voxft$ is fed to a language feature extractor.
This head processes each voxel feature to output an embedding vector aligned to vision-language representations,
such as CLIP~\cite{radford2021learning}, aiming to inherit their open-vocabulary abilities.
This allows us to address the limitations of closed-vocabulary predictions encountered in supervised 3D occupancy prediction models, which are bound to a set of predefined visual classes. In contrast, our representation enables us to perform 3D language-driven tasks such as zero-shot 3D semantic segmentation. 
Similarly to the occupancy head, the 3D-language head consists of $N_\text{ft}$ blocks with configuration \texttt{Linear-Softplus-Linear}, where each linear layer outputs $C_\text{ft}^\text{hidden}$ features, and a final linear layer that outputs $C_\text{ft}^\text{out}$-dimensional vision language embedding for 
each voxel. It outputs the tensor
\begin{equation}
    \Occ_\text{ft} = \fthead \left( \voxft \right)  \in \mathbb{R} ^ {H_\text{V}\times W_\text{V}\times D_\text{V}\times C_\text{ft}^\text{out}},
\end{equation}
containing the predicted vision-language embedding of each voxel.

\subsection{Tri-modal self-supervised training}
\label{sec:trimodal_learning}

The goal is to train the network architecture described in \autoref{sec:architecture} to predict the 3D occupancy map together with language-aware features for each occupied voxel. In turn, this will enable 3D open-vocabulary tasks such as 3D zero-shot segmentation or language-driven search. The main challenge is obtaining the appropriate 3D-grounded language annotations, which is expensive to do manually. Instead, we propose a tri-modal self-supervised learning algorithm that leverages three modalities: (i) images, (ii) language, and (iii) LiDAR point clouds. 
Specifically, we employ a pre-trained image-language network to generate image-language features for the input images. These features are then mapped to the 3D space using registered LiDAR point clouds, resulting in 3D grounded image-language features. These grounded features serve as training targets for the network.
The training algorithm is illustrated in \autoref{fig:proposed_approach}(b). The training is implemented via two losses used to train the two heads of the proposed architectures
jointly with the 2D-to-3D encoder. The details are given next.

\textbf{Occupancy loss.~} 
We guide the occupancy head~$\occhead$ to perform a class-agnostic occupancy prediction by the available unlabeled LiDAR point clouds, which we convert to occupancy prediction targets~$T_\text{occ}\in\{0,1\}$.
Each voxel location $x$ containing at least one LiDAR point is labeled as `occupied' (i.e., $T_\text{occ}(x)=1$) 
and as `empty' otherwise ($T_\text{occ}(x)=0$).
Having these targets, we supervise the occupancy prediction head densely at all locations of the voxel grid. The occupancy loss~$\mathcal{L}_\text{occ}$ is a combination of cross-entropy loss $\mathcal{L}_\text{CE}$ and Lov\'asz-softmax~\cite{berman2018lovasz} loss~$\mathcal{L}_\text{Lov}$:
\begin{equation}
    \mathcal{L}_\text{occ} \left( \Occ_\text{occ}, \Targ_\text{occ} \right) = \mathcal{L}_\text{CE} \left( \Occ_\text{occ}, \Targ_\text{occ} \right) + \mathcal{L}_\text{Lov} \left( \Occ_\text{occ}, \Targ_\text{occ} \right),
\label{eq:occ_loss}
\end{equation}
where $\Occ_\text{occ}$ is the predicted occupancy tensor and  $\Targ_\text{occ}$ the tensor of corresponding occupancy targets.

\textbf{Image-language distillation.~} 
Unlike the occupancy prediction head that is supervised densely at the level of voxels, we supervise the 3D-language head at the level of points $p_n \in P_\text{cam}$ which project to at least one of the cameras, i.e., $P_\text{cam} \subset P$, where $P$ is the complete point cloud. This is required to obtain feature targets from the language-image pre-trained model $\imext$.

To get a feature target for a 3D point $p_n\in P_\text{cam}$ in the voxel feature grid, we use the known camera projection function $\Pi_c$ that projects 3D point $p_n$ into 2D point $u_n= (u_n^{(x)},u_n^{(y)})$, where $(u_n^{(x)},u_n^{(y)})$ are $(x,y)$ coordinates of point $u_n$
in camera $c$:
\begin{equation}
u_n=\Pi_c\left( p_n \right).
\label{eq:projection}
\end{equation}
This way, we get a set of 2D points $U=\{\Pi_c\left( p_n \right)\}_{n=1}^{N}$ in the camera coordinates.To obtain feature targets $\Targ_\text{ft}$ for 3D points in $P_\text{cam}$ with corresponding 2D projections $U$ in camera $c$, we run the language-image-aligned feature extractor $\imext$ on image $\mathbf{I}_c$, and use the 2D projections' coordinates to sample from the resulting feature map, i.e.,
\begin{equation}
 \mathrm{T}_\text{ft}=\left\{ \imext \left( \mathbf{I}_c\right) [ u_n^{(x)},u_n^{(y)} ] \right\}_{n=1}^N \in \mathbb{R}^{N\times C_\text{ft}^\text{out}},
\end{equation} where $[x,y]$ is an indexing operator in the extracted feature map.

To train the 3D language head, we use $L_2$ mean squared error loss between the targets $\Targ_\text{ft}$ and the predicted features $\tilde{\Occ}_\text{ft} \in \mathbb{R}^{N\times C_\text{ft}^\text{out}}$ computed from $h$ for the 3D point locations in $P_\text{cam}$: %
\begin{equation}
    \mathcal{L}_\text{ft} = \frac{1}{N C_\text{ft}^\text{out}} \|\Targ_\text{ft} - \tilde{\Occ}_\text{ft}\|^2,
\end{equation}
where $\| \cdot \|$ is the Frobenius norm.

\textbf{Final loss.~} 
The final loss used to train the whole network is a weighted sum of the \textit{occupancy} and \textit{image-language} losses. We use a single hyperparameter $\lambda$ to balance the weighting of the two losses:
\begin{equation}
    \mathcal{L} = \mathcal{L}_\text{occ} + \lambda \mathcal{L}_\text{ft}.
    \label{eq:final_loss}
\end{equation}

\subsection{3D open-vocabulary test-time inference}

Once trained, as described in~\autoref{sec:trimodal_learning}, our model supports different 3D open-vocabulary tasks at test time. We focus on the following two: (i) zero-shot 3D semantic segmentation and (ii) language-driven 3D grounding. 

\textbf{Zero-shot 3D semantic segmentation from images.~}
Given an input test image, the 3D-text-aligned voxel features produced by our model support zero-shot 3D segmentation for a target set of classes specified via input text queries (prompts), as illustrated in~\autoref{fig:teaser}. Unlike supervised approaches that necessitate retraining when the set of target classes changes, our approach requires training the model only once. We can effortlessly adjust the number of segmented classes by providing a different set of input text queries.
In detail, at test time, we proceeded along the following steps. 
First, a set of test surround-view images $\mathbf{I}$ from one location is fed into the trained \ours~network, resulting in class-agnostic occupancy prediction $\Occ_\text{occ}$ via the occupancy head $\occhead$, and language-aligned feature predictions $\Occ_\text{ft}$ via the 3D-language head $\fthead$. 
Next, as described in~\cite{gu2021open}, we generate a set of query sentences for each text query using predefined templates. These queries are input into the pre-trained language-image encoder $f_\text{text}$, resulting in a set of language features. We compute the average of these features to obtain a single text feature per query.
Finally, considering $M$ such averaged text features, one for each of the $M$ target segmentation classes, we measure their similarity to the predicted language-aligned features $\Occ_\text{ft}$ at occupied voxels obtained from $\Occ_\text{occ}$. We assign the label with the highest similarity to each occupied voxel.

\textbf{Language-driven 3D grounding.~}
The task of language-driven 3D grounding is performed in a similar manner. However, here, only a single input language query is given. Once determining the occupied voxels from $\Occ_\text{occ}$, we compute the similarity between the input text query encoded via the language-image encoder $f_\text{text}$ and predicted language-aligned features $\Occ_\text{ft}$ at the occupied voxels. The resulting similarity score can be visualized as a heat map, as shown in \autoref{fig:teaser}, or thresholded to obtain the location of the target query.
\section{Experiments}
\label{sec:experiments}

This section studies architectural design choices and demonstrates the capabilities of the proposed approach. First, in~\autoref{sec:experimental_setup}, we describe the experimental setup used, particularly the dataset, metrics, proposed evaluation protocol, and implementation details.
Then, we compare our model to the state of the art in~\autoref{sec:comparison}.
Next, we present studies on training hyperparameter sensitivity in~\autoref{sec:ablations} and finally show qualitative results in~\autoref{sec:qualitative}.

\subsection{Experimental setup}
\label{sec:experimental_setup}

We test the proposed approach on autonomous driving data, which provides a challenging test bed. %

\textbf{Dataset.~}
We use the nuScenes~\cite{caesar2020nuscenes} dataset composed of 1000
sequences in total, divided into 700/150/150 scenes for train/val/test splits. Each sequence consists of $30-40$ scenes resulting in $28,130$ training and in $6,019$ validation scenes.
The dataset provides 3D point clouds captured with 32-beam LiDAR, surround-view images obtained from six cameras mounted at the top of the car, and projection matrices between the 3D point cloud and cameras.
LiDAR point clouds are annotated with 16 semantic labels.
When using subsets of the complete dataset for ablations, we sort the scenes by their timestamp and take every $N$-th scene, e.g., every second scene in the case of a $50$\% subset.

\textbf{Metrics.~}
To evaluate our models on the task of \task, we need to convert the point-level semantic annotations from LiDAR to voxel-level annotations. We do this by taking the most-present label inside each voxel.
As we aim at semantic segmentation, our main metric is mean Intersection over Union (mIoU), which we use in the evaluation protocol proposed in the next paragraph. Additionally, we measure the class-agnostic occupancy Intersection over Union (IoU).
\crd{
For the retrieval benchmark, we report the average precision (AP) for each query, 
\PP{the mean of which overall queries yield the mean average precision (mAP)}.
}

\crd{
\textbf{New benchmark for open-vocabulary language-driven 3D retrieval.~}
To evaluate the retrieval capabilities, 
we collected a new \emph{language-driven 3D grounding \& retrieval benchmark equipped with natural language queries}. To build this benchmark, we annotated 3D scenes from various splits of the nuScenes dataset with ground-truth spatial localization for a set of natural language open-vocabulary queries. 
The resulting set contains 105 samples in total, which are divided into 42/28/35 samples from train/val/test splits of the nuScenes dataset.
Given the query, the objective is to retrieve all relevant 3D points from the LiDAR point cloud. Results are evaluated using the precision-recall curve; negative data are all the non-relevant 3D points in the given scene.
For evaluation purposes, we report numbers on a concatenated set consisting of samples from the validation and test splits (63 samples).
To annotate the 3D retrieval ground truth, we (1) manually provide the bounding box of the relevant object(s) in the image domain, (2) use Segment Anything Model~\cite{Kirillov_2023_ICCV} guided by our manual bounding box to 
produce a binary mask of this object, (3) project the LiDAR point cloud into the image, and (4) assign each 3D point a label corresponding to its projection into the binary mask.
Furthermore, we use HDBSCAN~\cite{mcinnes2017hdbscan} to filter points that are projected to the mask in the image but, in fact, do not belong to the object. 
This resolves the imprecisions caused by projection.
}

\textbf{New evaluation protocol for \task.} The relatively new task of \ task~ has no established evaluation protocol yet. TPVFormer~\cite{huang2023tri} did not introduce any 
evaluation protocol and provided only qualitative results.
Having semantic labels only from LiDAR points, i.e., not in the target voxel space, makes it challenging to evaluate. 
Since voxel semantic segmentation consists of both \textit{occupancy prediction} of the voxel grid and \textit{classification of occupied voxels}, 
it is not enough to evaluate just at the points of ground-truth information from the LiDAR, as this does not consider free space prediction. 
\begin{wrapfigure}[12]{r}{0.45\textwidth} 
   \centering
        \resizebox{0.45\columnwidth}{!}{%
        \includegraphics[width=0.5\linewidth]{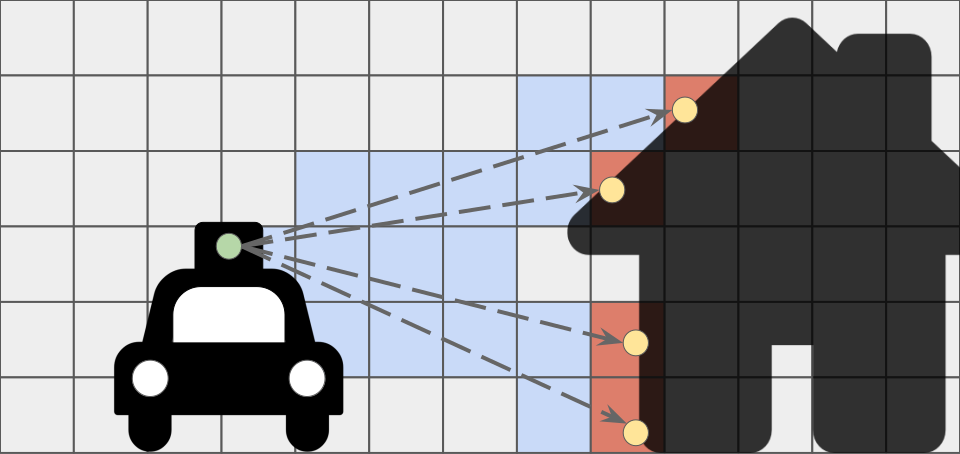}
        }
    \caption{\textbf{Validation labels}: 
    \textcolor{free}{blue~=~\textit{free}}, 
    \textcolor{red!80}{red~=~\textit{occupied}},
    and \textcolor{gray!120}{gray~=~\textit{ignored}} voxels.}
    \label{fig:val_labels}
\end{wrapfigure} 
To tackle this, we take inspiration from~\cite{boulch2022also} and propose to obtain the evaluation labels from the available LiDAR point clouds, as depicted in~\autoref{fig:val_labels} and described next.
First, LiDAR rays passing through 3D space set the labels of intersected voxels to \emph{free}.
Second, voxels containing LiDAR points are assigned the most frequent semantic label of points lying within (or an \textit{occupied label} in the case of class-agnostic evaluation). 
Third, all other voxels are \textit{ignored} during evaluation, as any LiDAR ray did not observe them, and we are not sure whether they are occupied or not.

\textbf{Implementation details.~}
We use the recent TPVFormer~\cite{huang2023tri} as the backbone for the 2D-3D encoder. It takes surround-view images on the input and produces a voxel grid of size $100\times100\times8$, which corresponds to 
the volume 
$[-51.2\text{m},+51.2\text{m}]\!\times\!
[-51.2\text{m},+51.2\text{m}]\!\times\!
[-5\text{m},+3\text{m}]$ 
around the car.
For the language-image feature extractor, we use MaskCLIP +~\cite{zhou2022maskclip}, which provides features of dimension $C_\text{ft}^\text{out}=512$.
If not mentioned otherwise, we use the default learning rate of $2\text{e-}4$, Adam~\cite{DBLP:journals/corr/KingmaB14} optimizer, and a cosine learning rate scheduler with final learning rate $1\text{e-}6$, and with linear warmup from $1\text{e-}5$ learning rate for the first $500$ iterations. We train our models on 8$\times$A100 GPUs.
We use ResNet-101 as the image backbone in the $\voxext$ encoder and full-scale images on the input.
Both prediction heads have two layers, i.e., $N_\text{occ}=N_\text{ft}=2$, and %
$C_\text{occ}=512$ and $C_\text{ft}=1024$ 
feature channels.
With this architecture setup, we train our model on 100\% of the nuScenes training data for 12 epochs.
We put the same weight to the occupancy and feature losses, i.e., we set $\lambda=1$ in~\autoref{eq:final_loss}. We ablate these choices in~\autoref{sec:ablations}.

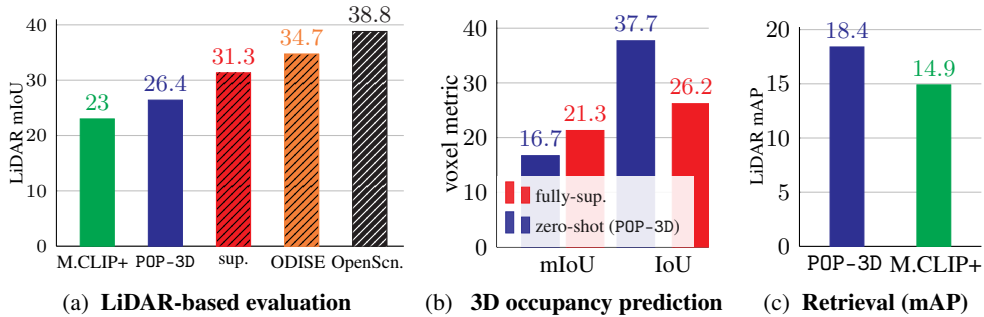
\begin{figure*}[ht!]
    \subfloat[
    \textbf{LiDAR-based evaluation}
    \label{fig:ablation_gt_occupancy}
    ]{
       \pgfplotstableread[row sep=\\,col sep=&]{
    model & result \\
    MaskCLIP+    & 23.0 \\
    \ours        & 26.4 \\
    fully-sup.   & 31.3 \\
    ODISE        & 34.7 \\
    OpenScene    & 38.8 \\
    }\mydata

  \begin{tikzpicture}
    \begin{axis}[ybar=-1cm,
			axis x line*=bottom,
			axis y line*=left,
			height=4.6cm,
			width=6.2cm,
                font=\footnotesize,
                title style={font=\footnotesize},
                label style={font=\scriptsize},
                tick label style={font=\scriptsize},
			bar width=0.5cm,
                ymajorgrids,
			ylabel={LiDAR mIoU},
                y label style={at={(-0.07,0.5)}},
                tickwidth = 0pt,
                ymin = 0,
                ymax = 41,
                xmin=0.4,
                xmax=5.5,
                xtick={0.93,2,3,3.95,4.97},
                xticklabels={M.CLIP+,\ours,sup.,ODISE,OpenScn.},
			nodes near coords,
			nodes near coords align={vertical},
                every axis plot/.append style={
                  ybar,
                  bar width=.5,
                  bar shift=0pt,
                  fill
                }
            ]
			\addplot[fill, Green] coordinates {(1,23.0)};
			\addplot[fill, Blue] coordinates {(2,26.4)};
			\addplot[fill, Red, postaction={pattern=north east lines}] coordinates {(3,31.3)};
                \addplot[fill, Orange, postaction={pattern=north east lines}] coordinates {(4,34.7)};
                \addplot[fill, Black, postaction={pattern=north east lines,pattern color=White}] coordinates {(5,38.8)};
    \end{axis}
\end{tikzpicture}
    }
    \subfloat[
    \textbf{
    3D occupancy prediction
    } 
    \label{fig:final_training}
    ]{
         \begin{tikzpicture}
    \begin{axis}[ybar=-1cm,
			axis x line*=bottom,
			axis y line*=left,
			height=4.6cm,
			width=4.4cm,
                font=\footnotesize,
                title style={font=\footnotesize},
                label style={font=\footnotesize},
                tick label style={font=\footnotesize},
			bar width=0.5cm,
			ylabel={voxel metric},
                y label style={at={(-0.14,0.5)}},
                ymajorgrids,
                ymin = 0,
                xtick = data,
                xticklabels={,mIoU,IoU,},
                tickwidth = 0pt,
                xtick={0.5,0.9,1.5},
                xmin=0.5,
                xmax=1.7,
			nodes near coords,
			nodes near coords align={vertical},
                legend cell align={left},
                legend style={at={(0.,0.)},anchor=south west,fill opacity=0.9, draw=none,font=\scriptsize},
   ]
			\addplot[fill, white, opacity=0] coordinates {(0.5,16.7) (2.,37.7)};
			\addplot[fill, Red] coordinates {(1,21.3) (1.6,26.2)};
			\addplot[fill, Blue] coordinates {(0.96,16.7) (1.5,37.7)};
                \legend{,fully-sup., zero-shot (\ours)};
    \end{axis}
\end{tikzpicture}
    }
    \subfloat[
    \textbf{\crd{Retrieval (mAP)}} 
    \label{fig:retrieval}
    ]{
         \begin{tikzpicture}
    \begin{axis}[ybar=-1cm,
			axis x line*=bottom,
			axis y line*=left,
			height=4.5cm,
			width=4.cm,
                font=\footnotesize,
                title style={font=\footnotesize},
                label style={font=\scriptsize},
                tick label style={font=\footnotesize},
			bar width=0.45cm,
			ylabel={LiDAR mAP},
                ymajorgrids,
                ymin = 0,
                xtick = data,
                xticklabels={\ours,M.CLIP+},
                tickwidth = 0pt,
                xtick={1,3},
                xmin=0,
                xmax=4,
			nodes near coords,
			nodes near coords align={vertical},
                y label style={at={(-0.15,0.5)}},
   ]
                \addplot[fill, Green] coordinates {(2.5,14.9)};
			\addplot[fill, Blue] coordinates {(1.5,18.4)};
    \end{axis}
\end{tikzpicture}
    }
    \caption{
    \textbf{Comparison to the state of the art}. We compare our \ours~approach to different baselines using  (a) the LiDAR-based evaluation, (b) occupancy evaluation\crd{, and (c) open-vocabulary language-driven retrieval}.
    In (a), our zero-shot approach~\ours~outperforms the strong MaskCLIP+~\cite{zhou2022maskclip} (M.CLIP+) baseline while closing the gap to the fully supervised.
    \crd{
Other recent methods using supervision and requiring LiDAR points during inference
    (ODISE~\cite{xu2023open} and OpenScene~\cite{Peng2023OpenScene}) are even better.
    } All methods that  
    require manual annotations during training are denoted by striped bars).
    In (b), our zero-shot approach \ours~surpasses the fully-supervised model~\cite{huang2023tri} on occupancy prediction (IoU) while reaching $78\%$ of its performance on semantic occupancy segmentation (mIoU).
    \crd{
    Finally, in (c), we present results of open-vocabulary language-driven retrieval on our newly composed dataset, where we compare our approach to the MaskCLIP+ baseline. We measure mAP on manually annotated LiDAR 3D points in the scene. Our~\ours~outperforms the MaskCLIP+ approach on this task by $3.5$ mAP points.}
    }
    \label{fig:my_label}
\end{figure*}

\subsection{Comparison to the state of the art}
\label{sec:comparison}

Here we compare our approach to four relevant methods: (i) the fully supervised (closed-vocabulary) TPVFormer~\cite{huang2023tri} and   the following three open-vocabulary image-based methods, namely (ii) MaskCLIP+~\cite{zhou2022maskclip}, \crd{
(iii) ODISE~\cite{xu2023open}, and (iv) OpenScene~\cite{Peng2023OpenScene}, which require 3D LiDAR point clouds on the input during the inference.
Please note that compared to methods (ii)-(iv), our~\ours~does not require (1) strong manual annotations (either in the image or point cloud domain) or (2) having point clouds on the input during the inference. Details are given next.
}

\textbf{Comparison to a fully-supervised TPVFormer~\cite{huang2023tri}.}
In figure \autoref{fig:final_training},
we compare our results to the supervised TPVFormer~\cite{huang2023tri} in terms of class-agnostic IoU and (16+1)-class mIoU (16 semantic classes plus the \textit{empty} class) on the nuScenes~\cite{caesar2020nuscenes} validation set. Interestingly, our %
model outperforms its supervised counterpart in the class-agnostic IoU by $11.5$ points, showing superiority in the prediction of the occupied space. This can be attributed to different training schemes of the two methods: in the fully-supervised case, the \textit{empty} class competes with the other semantic classes, whereas in our case the occupancy head performs only class-agnostic occupancy prediction. Next, for the (16+1)-class semantic occupancy segmentation, we can see that our zero-shot approach reaches $\approx78\%$ 
of the supervised counterpart performance, which we consider a strong result given that the latter requires manually annotated point clouds for training. %
In contrast, our approach is zero-shot and does not require any manual point cloud annotations at training. %
These results pave the way for language-driven vision-only 3D occupancy prediction and semantic segmentation in automotive applications. We show qualitative results of our~\ours~approach in~\autoref{fig:qual_v2} and in the supplementary materials.

\textbf{Comparison to MaskCLIP+~\cite{zhou2022maskclip}.}
In 
\autoref{fig:ablation_gt_occupancy} we compare the quality of the 3D vision-language features learnt by our
\ours ~ approach against the strong \textit{MaskCLIP+}\cite{zhou2022maskclip} baseline. 
In detail, we project the 3D LiDAR points to the 2D image(s) space, sample MaskCLIP+\cite{zhou2022maskclip} features extracted from the 2D image at the projected locations and backproject those extracted features back to 3D via the LiDAR rays. 
Note that \textit{MaskCLIP+} features are used in our tri-modal training to represent the language modality. Hence, it is interesting to evaluate the benefits of our approach in comparison to directly transferring \textit{MaskCLIP+} features to 3D. 
For a fair comparison, we evaluate only the LiDAR points with a projection to the camera, i.e., this evaluation considers only the classification of the 3D points, not the occupancy prediction itself.
We call this metric LiDAR mIoU. 
Our \ours ~outperforms MaskCLIP+ ($26.4$ vs. $23.0$ mIoU), i.e., our method learns better 3D vision-language features than its teacher while also not requiring LiDAR data at test time (as MaskCLIP+ does). Finally, 
\autoref{fig:ablation_gt_occupancy} shows that \ours~reaches $\approx84\%$  of the performance of the fully-supervised model~\cite{huang2023tri}.

\crd{
\textbf{Comparison to open-vocabulary methods that require additional supervision.}
Furthermore, we compare our approach to ODISE~\cite{xu2023open} and OpenScene~\cite{Peng2023OpenScene}, which both require manual supervision during training.
ODISE requires panoptic segmentation annotations for training, while OpenScene uses features from either LSeg~\cite{li2022languagedriven} or OpenSeg~\cite{ghiasi2022scaling}, which are two image-language encoders that are trained with supervision from manually provided segmentation masks. 
We report results using OpenSeg. As 
\autoref{fig:final_training} shows these methods perform best, which can be attributed to additional manual annotations available during training.
}

\crd{
\textbf{Open-vocabulary language-driven retrieval.}
Given a text query of the searched object, the goal is to retrieve all 3D points belonging to the object in the given scene.
During the evaluation, to get the relevance of LiDAR points to the query text description, we follow the same approach as for the task of zero-shot semantic segmentation, i.e., we pass the images to our model, get features aligned with the text, and compute their relevance to the given text query. This gives a score for every 3D point in the scene. In the ideal case, the points belonging to the target object should have the highest score.
We compare our method with MaskCLIP+ and report results in \autoref{fig:retrieval}. Our approach exhibits superior mAP compared to MaskCLIP+, achieving $18.4$ mAP while MaskCLIP+ obtains mAP of $14.9$.
}

\subsection{Sensitivity analysis}
\label{sec:ablations}

Here we study the sensitivity of our model to various hyperparameters. 
Except otherwise stated, 
for this study we use half-resolution input images, i.e., $450\times800$, the ResNet-50 backbone, and train for 6 epochs using 50\% of the nuScenes training data.

\begin{table*}[ht!]    
\caption{\textbf{Sensitivity analysis.}
We investigate here the impact of loss weight $\lambda$ in the final loss function (a), the image resolution and image backbone (b), and the depth of the prediction heads (c). %
}
\hspace{4pt}
\begin{minipage}{0.15\textwidth}{
    \subfloat[
    \textbf{Loss weight $\lambda$ impact}
    \label{tab:ablation_loss}
    ]{
    \centering
    \begin{tabular}{l|cc} \toprule
    $\lambda$ & mIoU & IoU  \\
    \midrule
    $1.00$ & 12.0 & 30.0 \\
    $0.50$ & 12.0 & 30.5 \\
    $0.25$ & 11.9 & 30.5 \\
    \bottomrule
    \end{tabular}
    }
    }
\end{minipage}
\hspace{5em}
\begin{minipage}{0.25\textwidth}{
    \subfloat[
    \textbf{Image resolution and backbone}
    \label{tab:ablation_scaling}
    ]{
    \centering
    \begin{tabular}{l|cc} \toprule
    \multicolumn{1}{c|}{image} & \multicolumn{2}{c}{mIoU} \\
    \multicolumn{1}{c|}{resolution} & RN50 & RN101 \\
    \midrule
    450$\times$800 & 12.0 & 15.1 \\
    900$\times$1600 & 12.3 & 15.2 \\ \bottomrule
    \end{tabular}
    }
}
\end{minipage}
\hspace{5em}
\begin{minipage}{0.15\textwidth}{
    \subfloat[
    \textbf{Depth of prediction heads} 
    \label{tab:ablation_head}
    ]{
    \centering
    \begin{tabular}{c|cc} \toprule
    & \multicolumn{2}{c}{mIoU} \\
    $N_\text{occ}$~$/$~$N_\text{ft}$ & 2 & 3 \\
    \midrule
    2 & 15.4 & 15.3 \\
    3 & 15.3 & 15.5 \\
    \bottomrule
    \end{tabular}
    }
}
\end{minipage}
\vspace*{-0.1cm}
\end{table*}

\vspace*{-0.2cm}
\paragraph{Loss weight $\lambda$.}
In 
~\autoref{tab:ablation_loss} 
we study the sensitivity of our model to the loss weight $\lambda$ of $\mathcal{L}_{ft}$ in \autoref{eq:final_loss}.
We see that the model's performance is not sensitive to $\lambda$. By default, we use $\lambda=1$.

\vspace*{-0.2cm}
\paragraph{Input resolution and image backbone.}
In 
~\autoref{tab:ablation_scaling} 
we experiment with (a) using half (450$\times$800) or full (900$\times$1600) input images, and (b) using ResNet-50 (RN50) or ResNet-101 (RN101) for the image backbone. Following~\cite{huang2023tri}, RN50 is initialized from MoCov2~\cite{chen2020improved} weights and RN101 from FCOS3D~\cite{wang2021fcos3d} weights.
We see that it is better to use the RN101 backbone while the input resolution has a small impact (with full resolution being better).

\vspace*{-0.2cm}
\paragraph{Depth of prediction head.} %
In 
~\autoref{tab:ablation_head} 
we study the impact of the $N_\text{occ}$ and $N_\text{ft}$ hyperparameters that control the number of hidden layers on the occupancy prediction head $g$ and 3D language head $h$ respectively, using RN101 as a backbone. 
We see that the depth of the two prediction heads does not play a major role and it is slightly better to be the same, i.e, $N_\text{occ}=N_\text{ft}$.
Therefore, we opt to use $N_\text{occ}=N_\text{ft}=2$ in our experiments, as it performs well and requires less compute.

\subsection{Demonstration of open-vocabulary capabilities}
\label{sec:qualitative}

In \autoref{fig:qual_retrieval}, we provide visualizations of language-based 3D object retrievals inside a scene using text queries like ``building door'' and ``tire''.
For reference, green boxes denote the locations of reference objects (cars), 
to ease the orientation in the scene. 
The results show that our model can localize fine-grained language queries in 3D space.

\vspace*{-0.2cm}
\paragraph{Limitations.} First, given the voxel grid's low spatial resolution, our model does not discover small objects well. 
This is not a limitation of the method but of the currently used backbone architecture and input data. Second, another limitation is that our architecture does not natively support sequences of images as input, which might be beneficial for reasoning about semantic occupancy of occluded objects and areas appearing thanks to the relative motion of objects in the scene.
\section{Conclusion}

In this paper we propose \ours, a tri-modal self-supervised learning strategy with a novel architecture that enables open-vocabulary voxel segmentation from 2D images and, at the same time, improves the occupancy grid estimation by a significant margin over the state of the art. Our approach also outperforms the strong baseline of directly back-projecting 2D vision-language features into 3D via LiDAR and does not require LiDAR at test time. 
This work opens up the possibility of large-scale open-vocabulary 3D scene understanding driven by natural language.

\setlength{\tabcolsep}{0.5pt}
\begin{figure*}[ht!]
    \small
    \centering
    \includegraphics[width=4.5cm,height=3cm]{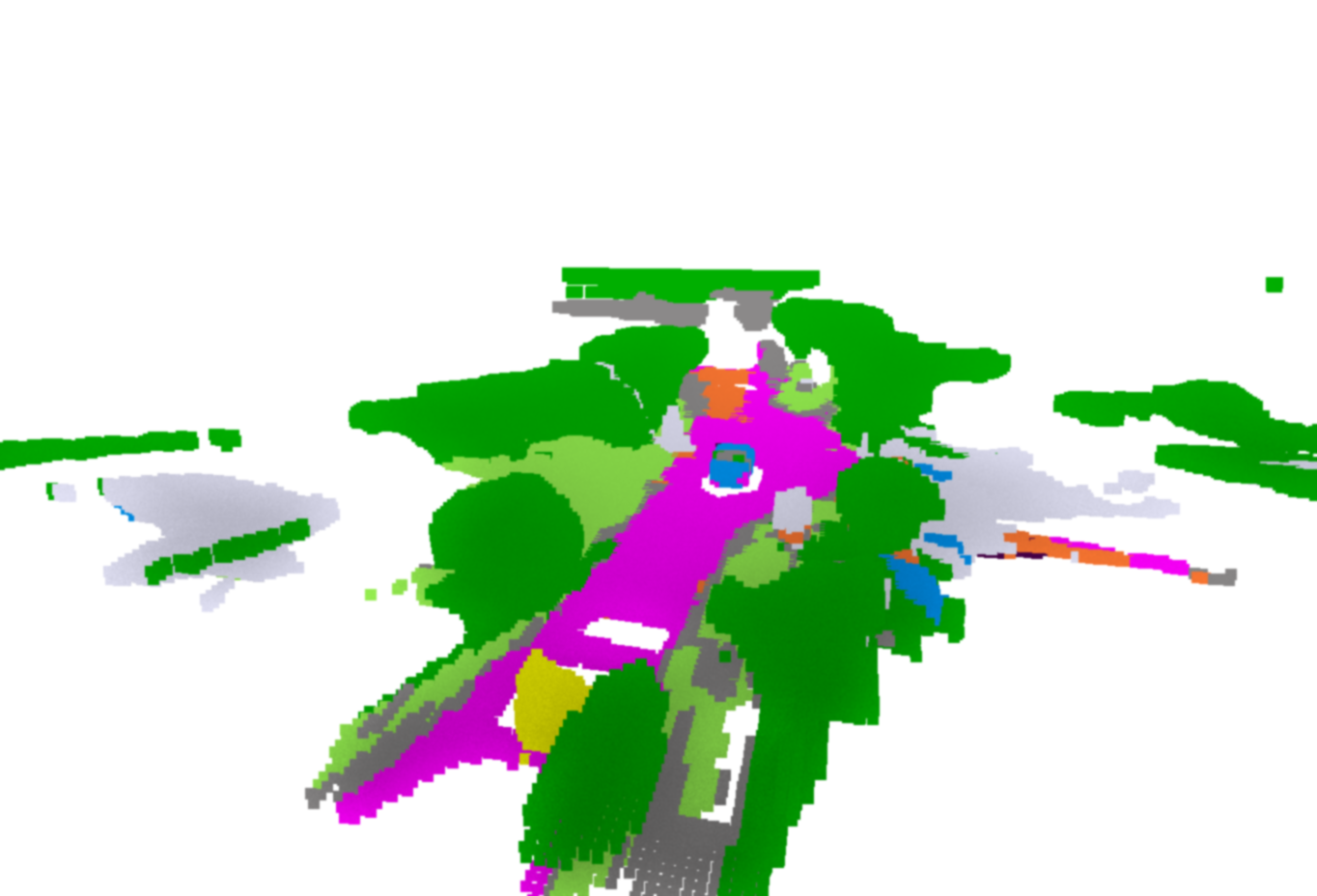}  
    \includegraphics[width=4.5cm,height=3cm]{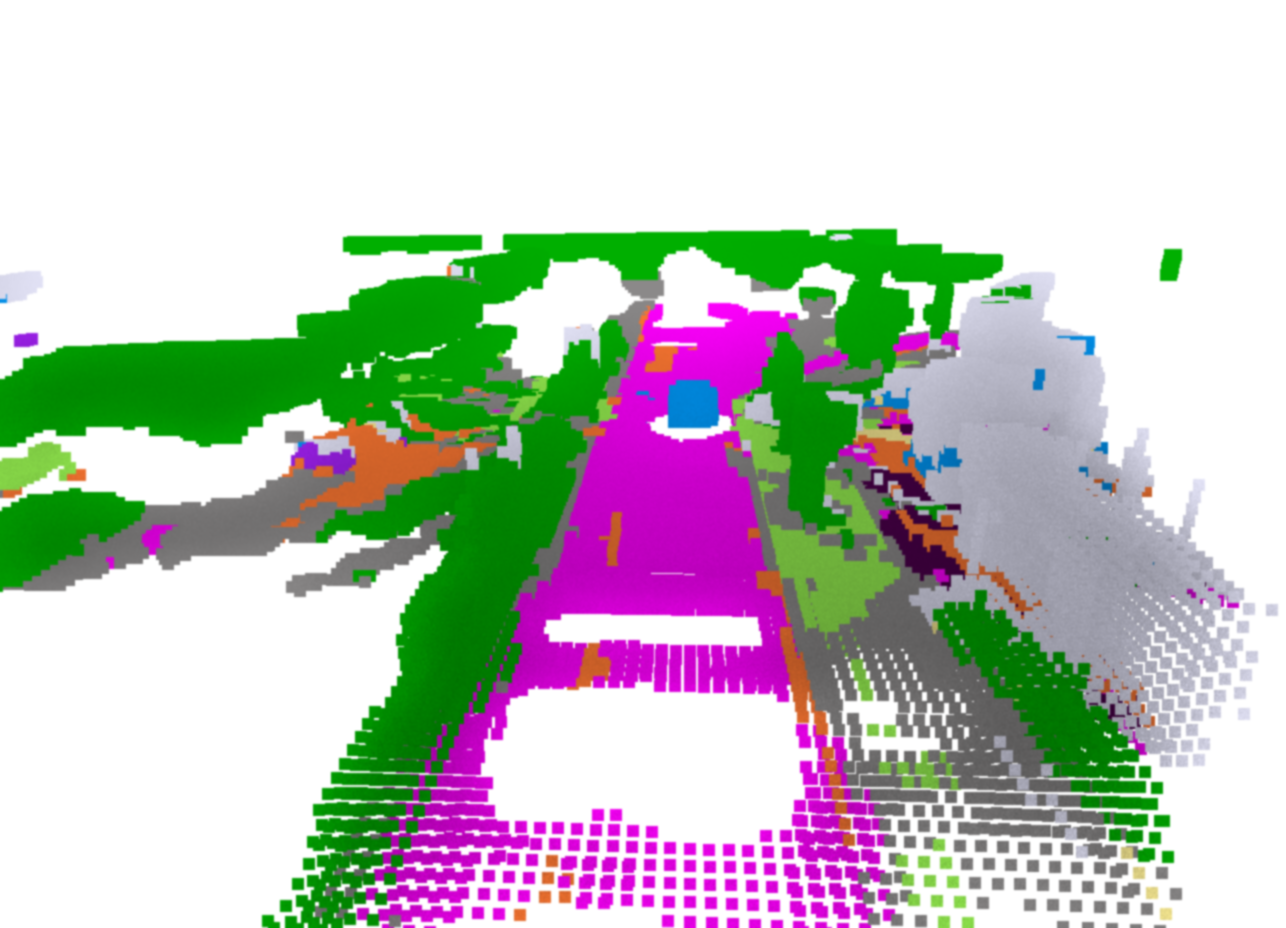} 
    \includegraphics[width=4.5cm,height=3cm]{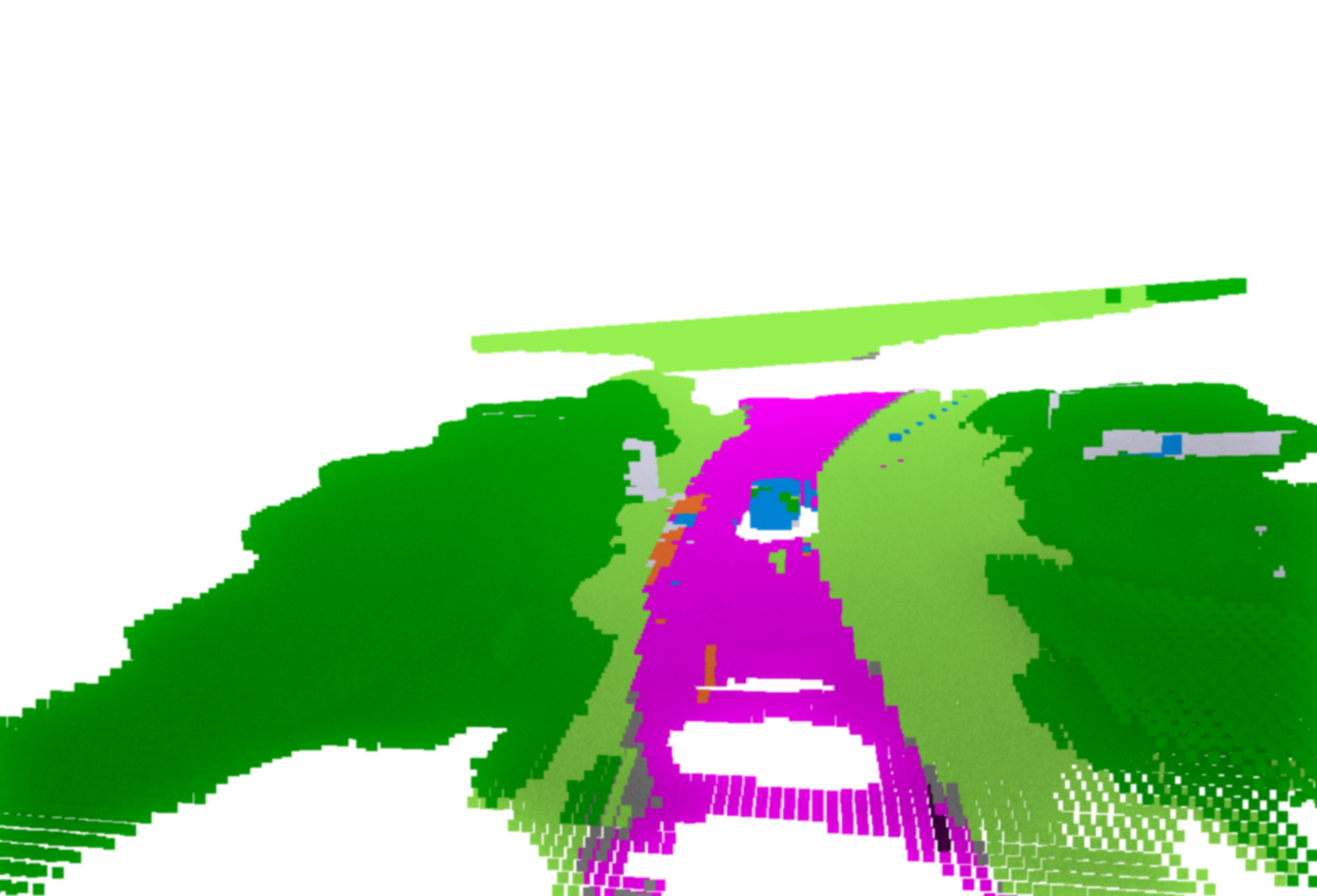} \\

    \includegraphics[width=4.5cm,height=3cm]{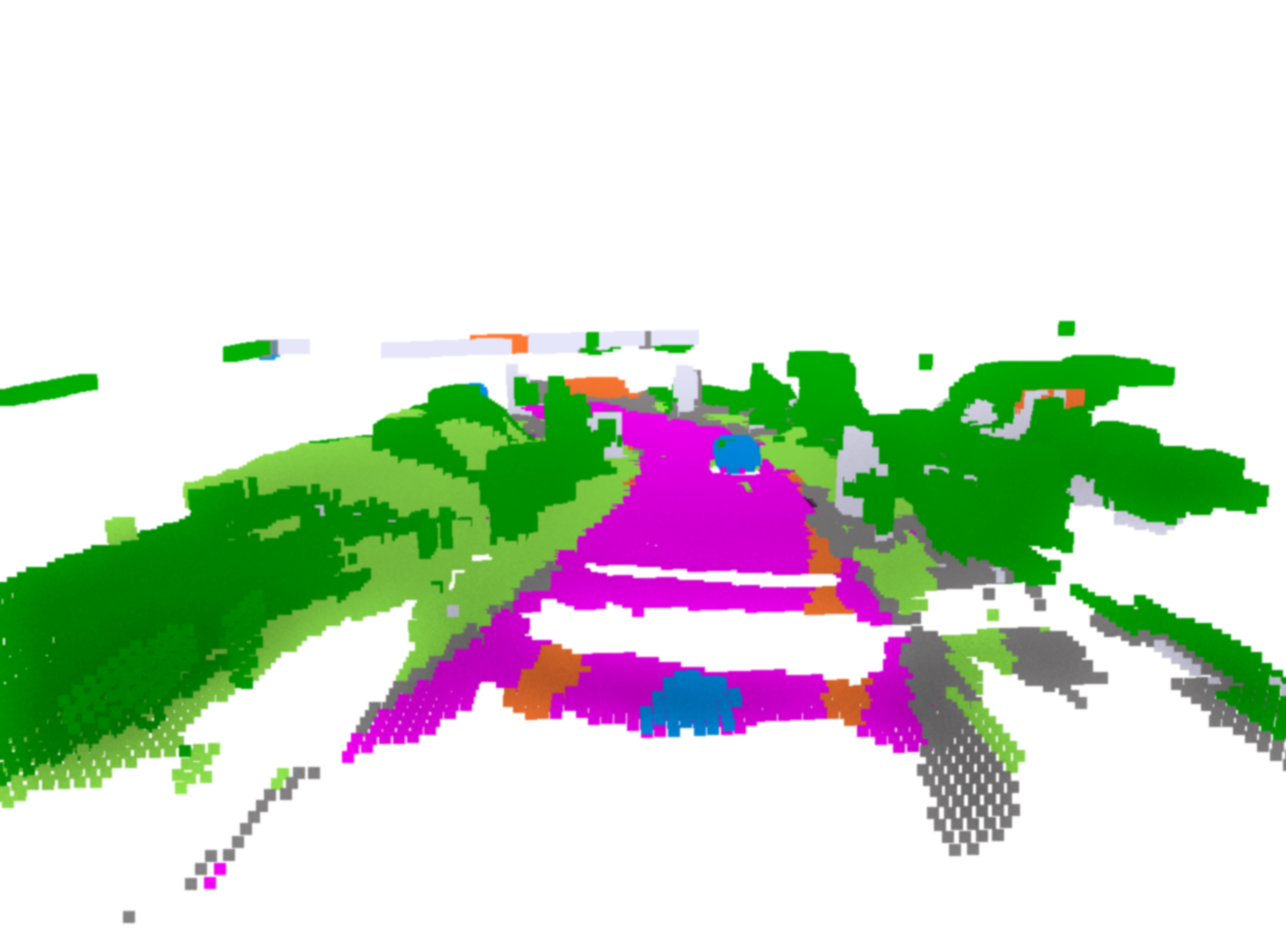} 
    \includegraphics[width=4.5cm,height=3cm]{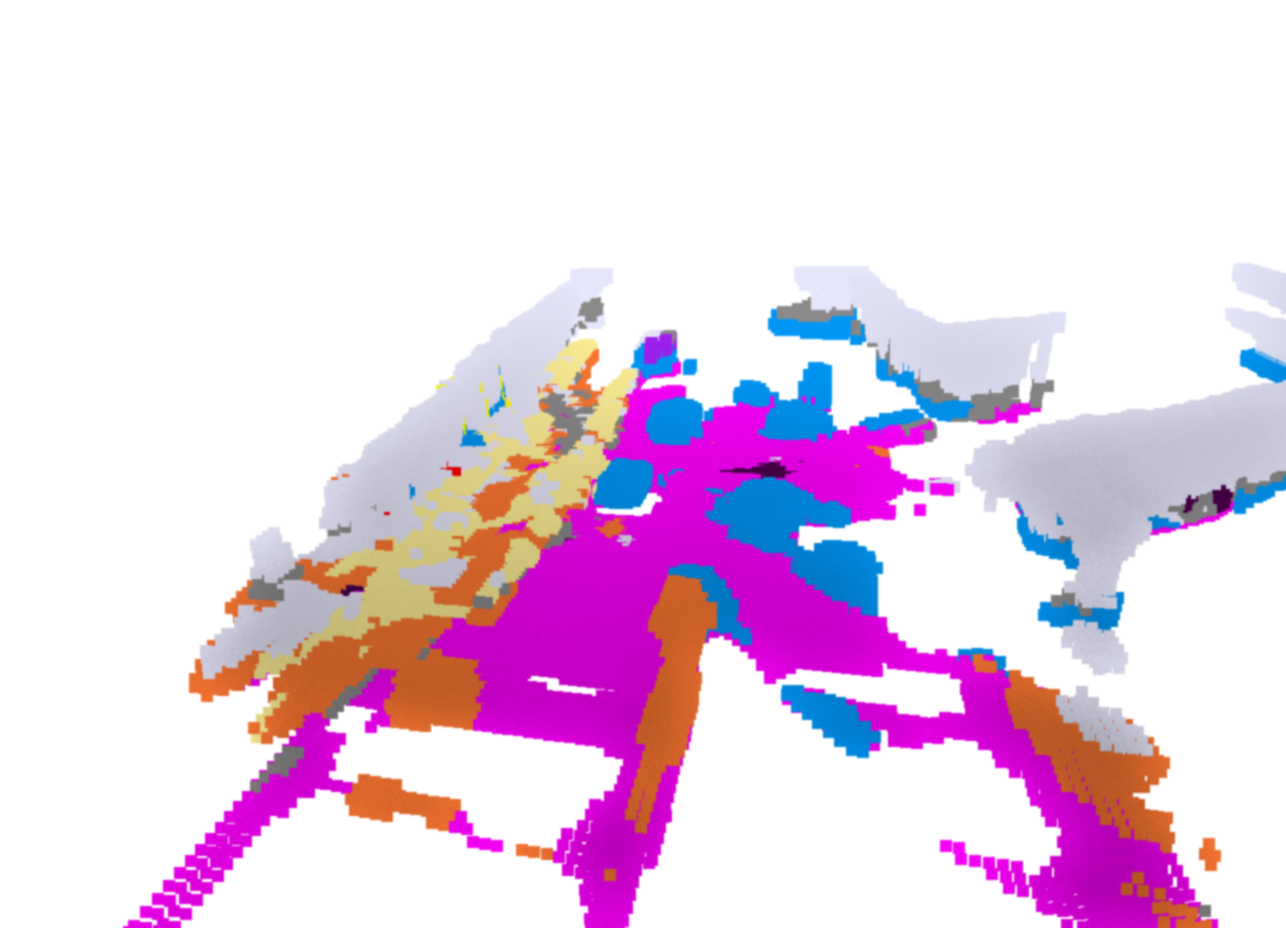} 
    \includegraphics[width=4.5cm,height=3cm]{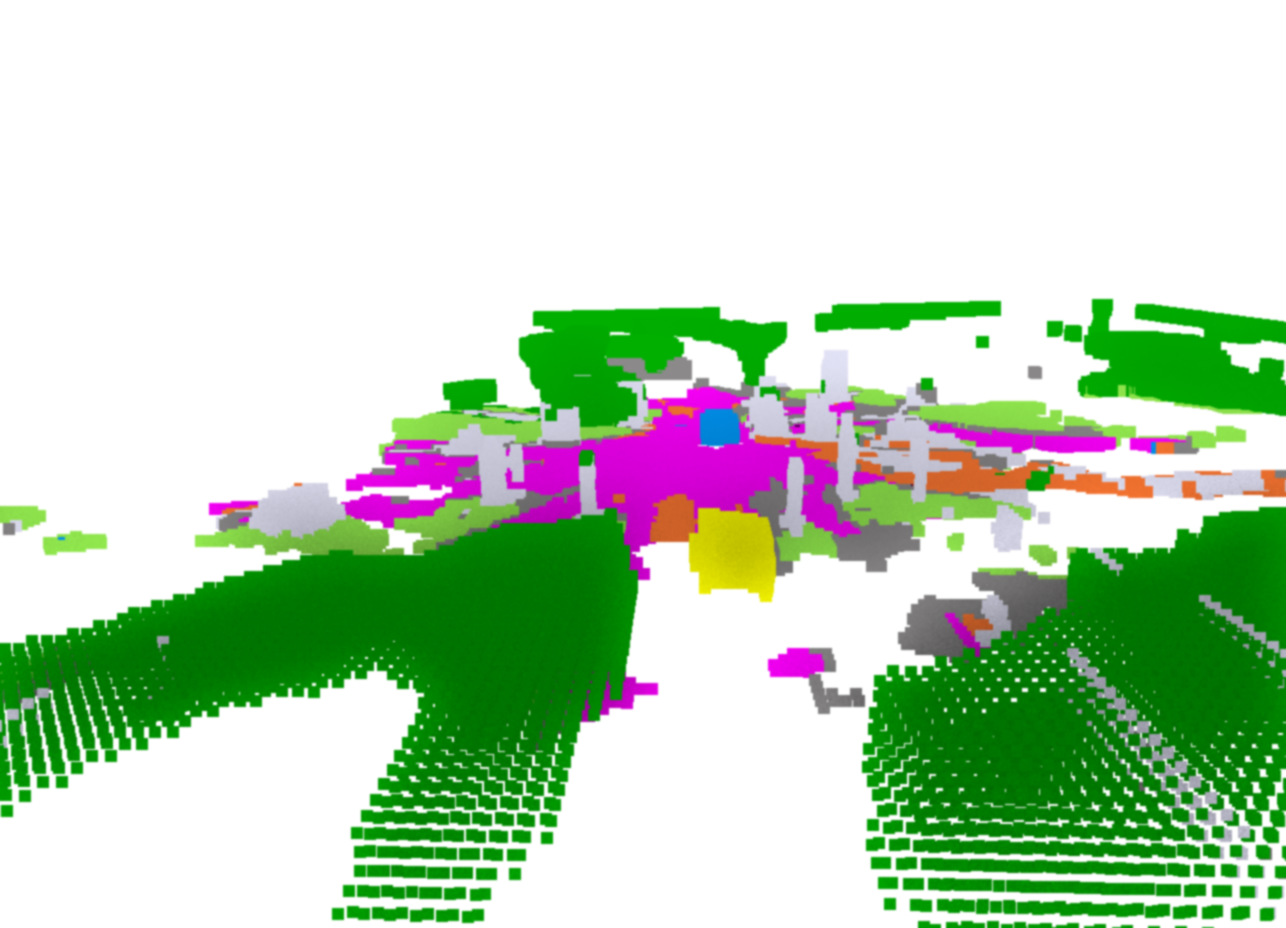} \vspace{-2pt} \\
    \caption{\textbf{Qualitative results of zero-shot semantic \task} on the 16 classes in the nuScenes~\cite{caesar2020nuscenes} validation split. Please note how our method is able to quite accurately localize and segment objects in 3D including road (magenta), vegetation (dark green), cars (blue), or buildings (gray) from only input 2D images and in a zero-shot manner, i.e. only by providing  natural language prompts for the target classes. Visualizations are shown on an interpolated 300x300x24 voxel grid.}
    \label{fig:qual_v2}
\end{figure*}

\begin{figure*}[ht!]
\centering
    \includegraphics[width=\linewidth]{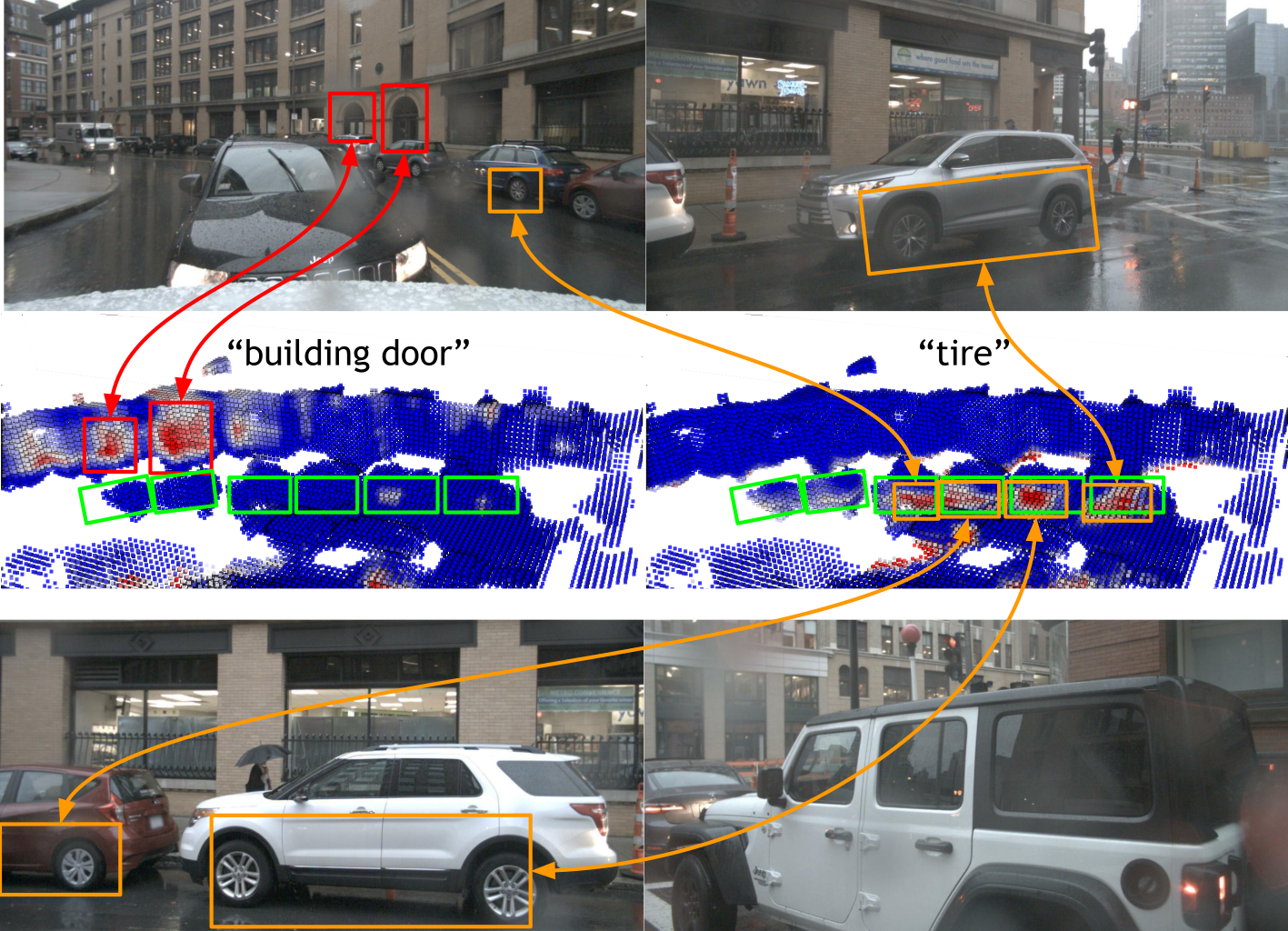}
    \caption{
    \textbf{Qualitative results showcasing the language-driven 3D grounding and retrieval.} 
    On the left (in red), we see the six input images passed to the \ours to get open-vocabulary 3D features (middle). Given the searched object name ("Black hatchback"), we compute the similarity with the 3D feature field and obtain a similarity heatmap (right).
    Language-based 3D retrieval is impossible using close-vocabulary methods such as~\cite{huang2023tri}.
     Please see additional results in the supplementary.}
     
    \label{fig:qual_retrieval}
\end{figure*}

\newpage\clearpage
{
\section*{Acknowledgements}
This work supported by the European Regional Development Fund under the project IMPACT (no. CZ.02.1.01\/0.0\/0.0\/15\_003\/0000468), and by CTU Student Grant SGS21\/184\/OHK3\/3T\/37.
This work was supported by the Ministry of Education, Youth and Sports of the Czech Republic through the e-INFRA CZ (ID:90254).
This research received the support of EXA4MIND, a European Union's Horizon Europe Research and Innovation programme under grant agreement N° 101092944. Views and opinions expressed are however those of the author(s) only and do not necessarily reflect those of the European Union or the European Commission. Neither the European Union nor the granting authority can be held responsible for them.
The authors have no competing interests to declare that are relevant to the content of this article.
Antonin Vobecky acknowledges travel support from ELISE (GA no 951847).
We acknowledge the support from Valeo.
}

{\small
\bibliographystyle{ieee_fullname}
\bibliography{references}
}

\newpage\clearpage
\appendix
\appendixpage

In this appendix, we first give additional details about the method in~\autoref{sec:supp_details}. \crd{Then, in~\autoref{sec:retrieval}, we provide additional details about the benchmark for open-vocabulary language-driven 3D retrieval.
Finally, }we present additional qualitative results in~\autoref{sec:supp_qual}.
\setcounter{footnote}{1}
\fullname{$^2$Czech Institute of Informatics, Robotics and Cybernetics at the Czech Technical University in Prague}
 
\section{Text queries for zero-shot \task}
\label{sec:supp_details}

This section investigates how selecting text queries assigned to specific ground-truth classes impacts semantic segmentation.

To simplify the analysis 
of the impact of the language prompt, we
study MaskCLIP+~\cite{zhou2022maskclip} features, which we also use as our training targets.
This choice allows us to uncover the capabilities associated with these features. Using the nuScenes~\cite{caesar2020nuscenes} dataset, we project the language-image-aligned features from MaskCLIP+ onto the corresponding LiDAR points. To measure the mIoU, we evaluate our approach on a subset comprising 25\% of the nuScenes validation set. It is important to note that, for a fair comparison, we calculate the mIoU only for the points with camera projections (other LiDAR points cannot have associated features from MaskCLIP+).

\crd{
\paragraph{Queries used for zero-shot semantic segmentation.} 
To get the text queries for the task of language-guided zero-shot semantic segmentation, we utilize the textual descriptions from the nuScenes~\cite{caesar2020nuscenes} dataset associated with every sub-class (names of the sub-classes are in the first column of~\autoref{tab:supp_descriptions}). 
We parse these descriptions into a set of queries (for every sub-class) and show them in the last column in~\autoref{tab:supp_descriptions}.
We do this for all the annotated classes in the dataset (second column).
}

\begin{table}[ht]
    \centering
    \renewcommand{\arraystretch}{1.1}
    \rowcolors{2}{gray!10}{white}
    \caption{\textbf{Queries used for zero-shot semantic segmentation.} We take the textual descriptions from the nuScenes~\cite{caesar2020nuscenes} dataset associated with every sub-class (names of the sub-classes are in the first column) and use them to get the individual queries associated with this class (the last column). We do this for all the annotated classes in the dataset.}
    \smallskip	
    \resizebox{0.8\linewidth}{!}{
    \begin{tabular}{llp{2.75in}} \toprule
Sub-class & Training class & Derived descriptions/queries \\
\midrule
noise & noise & ``any lidar return that does not correspond to a physical object, such as dust, vapor, noise, fog, raindrops, smoke and reflections''\\
adult pedestrian & pedestrian & adult\\
child pedestrian & pedestrian & child\\
construction worker & pedestrian & construction worker\\
personal mobility & ignore & skateboard; segway\\
police officer & pedestrian & police officer\\
stroller & ignore & stroller\\
wheelchair & ignore & wheelchair\\
barrier & barrier & ``temporary road barrier to redirect traffic''; concrete barrier; metal barrier; water barrier\\
debris & ignore & ``movable object that is left on the driveable surface''; tree branch; full trash bag\\
pushable pullable & ignore & ``object that a pedestrian may push or pull''; dolley; wheel barrow; garbage-bin; shopping cart\\
traffic cone & traffic cone & traffic cone\\
bicycle rack & ignore & ``area or device intended to park or secure the bicycles in a row''\\
bicycle & bicycle & bicycle\\
bendy bus & bus & bendy bus\\
rigid bus & bus & rigid bus\\
car & car & ``vehicle designed primarily for personal use''; car; vehicle; sedan; hatch-back; wagon; van; mini-van; SUV; jeep\\
construction vehicle & construction vehicle & vehicle designed for construction.; crane\\
ambulance vehicle & ignore & ambulance; ambulance vehicle\\
police vehicle & ignore & police vehicle; police car; police bicycle; police motorcycle\\
motorcycle & motorcycle & motorcycle; vespa; scooter\\
trailer & trailer & trailer; truck trailer; car trailer; bike trailer\\
truck & truck & ``vehicle primarily designed to haul cargo''; pick-up; lorry; truck; semi-tractor\\
driveable surface & driveable surface & ``paved surface that a car can drive''; ``unpaved surface that a car can drive''\\
other flat & other flat & traffic island; delimiter; rail track; stairs; lake; river\\
sidewalk & sidewalk & sidewalk; pedestrian walkway; bike path\\
terrain & terrain & grass; rolling hill; soil; sand; gravel\\
manmade & manmade & man-made structure; building; wall; guard rail; fence; pole; drainage; hydrant; flag; banner; street sign; electric circuit box; traffic light; parking meter; stairs\\
other static & ignore & ``points in the background that are not distinguishable, or objects that do not match any of the above labels''\\
vegetation & vegetation & bushes; bush; plants; plant; potted plant; tree; trees\\
ego vehicle & ignore & ``the vehicle on which the cameras, radar and lidar are mounted, that is sometimes visible at the bottom of the image''\\ \bottomrule
    \end{tabular}}
    \label{tab:supp_descriptions}
\end{table}
\renewcommand{\arraystretch}{1}

\paragraph{Limited-classes experiment.}
First, we conduct a controlled experiment with a limited set of five classes that are described by names `car,' `drivable surface,' `pedestrian,' `vegetation,' and `manmade' in the~nuScenes~\cite{caesar2020nuscenes} dataset. We refer to this specific setup as \texttt{original-5}, disregarding the other classes for the purpose of this study.
One can see that, for example, class name \textit{`manmade'} lacks descriptive specificity. In the text description of this class, we can find ``... buildings, walls, guard rails, fences, poles, street signs, traffic lights ...'' and more. We make similar observations for a number of class names in the nuScenes~\cite{caesar2020nuscenes} dataset. This observation highlights the limitation of relying solely on class names to guide text-based querying.

To study and address this limitation, we introduce two additional setups, namely \texttt{manmade-5} and \texttt{pedestrian-5}. %
In \texttt{manmade-5}, we replace the class name \textit{`manmade'} with \textit{`building'}, while in \texttt{pedestrian-5}, we use \textit{`person'} instead of \textit{`pedestrian'}. 
The results presented in the upper part of~\autoref{tab:ablation_queries} demonstrate the effectiveness of these changes. Specifically, replacing \textit{`manmade'} with \textit{`building'} improves the IoU for this category from 17.4 to 45.1, and using \textit{`person'} instead of \textit{`pedestrian'} increases the IoU from 1.3 to 14.6 for the respective class. These findings highlight the suboptimal use of original class names as text queries.

\begingroup
\setlength{\tabcolsep}{2.5pt}
\begin{table}[ht!]
   \caption{\textbf{Segmentation mIoU with a different number of target classes and text queries}. The first part of the table considers segmentation into 5 classes only, while the second part evaluates the complete set of 16 training classes. Results were obtained using 25\% of the validation split.}
   \label{tab:ablation_queries}
   \centering
   \smallskip
   \small
   \resizebox{0.5\linewidth}{!}{
   \begin{tabular}{l@{\hskip 1em} >{\columncolor[gray]{0.93}}c@{\hskip 1em}c@{\hskip 1em}c@{\hskip 1em}c@{\hskip 1em}c@{\hskip 1em}c} \toprule
   setup &  \multicolumn{6}{c}{visible points IoU} \\ \cmidrule(lr){2-7}
   \texttt{\{NAME\}-\{\#cls\}} & mIoU & car & road & ped. & veg. & man. \\
   \midrule
   \multicolumn{7}{l}{5 classes} \\
   \enspace\texttt{original-5} & 27.5 & 21.2 & 37.3 & 1.3 & 60.3 & 17.4 \\
   \enspace\texttt{manmade-5} & \textbf{34.7} & 28.1 & 37.3 & 1.6 & 61.2 & \textbf{45.1} \\
   \enspace\texttt{pedestrian-5} & \textbf{35.0} & 17.5 & 61.9 & \textbf{14.6} & 59.7 & 21.3 \\
   \midrule
   \multicolumn{7}{l}{16 classes} \\
   \enspace\texttt{original-16} & 10.2 & 25.8 & 0.9 & 3.0 & 51.3 & 0.5 \\
   \enspace\texttt{descriptions-16} & \textbf{23.0} & 37.9 & 57.5 & 16.9 & 62.6 & 45.4 \\
   \bottomrule
   \end{tabular}
   }
\end{table}
\endgroup

\paragraph{Training-classes experiment.}
Building upon these findings, we extend our study to include the full set of 16 classes used in the nuScenes dataset. We conduct experiments using two setups: i) \texttt{original-16}, which uses the original training class names from the nuScenes dataset, and 
ii) \texttt{descriptions-16}, where we utilize 
the detailed textual descriptions that are provided for each class in the nuScenes dataset (we explain in more detail this setup in the next paragraph).
By leveraging the textual descriptions provided by the nuScenes dataset, we can generate more informative %
and descriptive queries for each individual class, as demonstrated in~\autoref{tab:supp_descriptions}. This table presents the entire set of 32 (sub-)classes annotated in the nuScenes~\cite{caesar2020nuscenes} dataset, 
along with their mapping to the training classes and the corresponding derived queries.
The lower section of~\autoref{tab:ablation_queries} demonstrates the impact of modifying the text queries associated with individual classes in the nuScenes dataset. We observe that this simple adjustment significantly enhances the mIoU from $10.2$ to $23.0$, highlighting the significance of query selection. Based on these results, we have used
the \texttt{descriptions-16} setup for our experiments in the main paper.

The results suggest that further improvements could be achieved by 
carefully tuning the text queries. However, it is important to note that the focus of our paper is not on exploring query tuning; therefore, we do not delve further into this direction.

\paragraph{Using derived descriptions for segmentation.}
To utilize the derived queries presented in~\autoref{tab:supp_descriptions}, we begin by mapping the 32 sub-classes to the 16 training classes in the nuScenes dataset (note that some sub-classes are marked as `ignore' in the ``Training class'' column of~\autoref{tab:supp_descriptions} to indicate that they are actually ignored during evaluation).
For example, consider the training class `\textit{pedestrian}.' The sub-classes that are associated with this training class are: `\textit{adult pedestrian},' `\textit{child pedestrian},' `\textit{construction worker},' and `\textit{police officer}.'
We use the derived text descriptions (third column in~\autoref{tab:supp_descriptions}) of each of those sub-classes as text queries for the `\textit{pedestrian}' training class, resulting in the following set of queries: [\textit{adult, child, construction worker, police officer}]. 

This process produces a set of queries $Q$ with size $N^Q=|Q|=\sum_{c\in \{0\ldots 15\}} {N^Q_c}$, where $N^Q_c$ is number of queries associated with the training class $c\in \{0\ldots 15\}$.
Each query $q \in Q$ is mapped to a single training class $c$ via the mapping:
\begin{equation}
\mathcal{M}: \{0~\ldots ~N^Q-1\}\rightarrow\{0\ldots 15\}.   
\end{equation}

We follow a three-step process to assign a feature $o_\text{ft}$ from the set of all predicted features $O_\text{ft}$ to one of the training classes. First, we calculate the similarity between the feature $o_\text{ft}$ and each query. Next, we select the query with the highest similarity. Finally, we assign the corresponding training class label $c_\text{pred}$ based on the selected query. For example, if the query `\textit{police officer}' has the highest similarity, we assign the label `\textit{pedestrian}' to the feature $o_\text{ft}$. This can be formulated as:
\begin{equation}
    c_\text{pred} = \mathcal{M} \left( \argmax_{n\in \{0\ldots N^Q-1\}} { \left (\text{sim} \left( o_\text{ft}, q_n \right) \right) } \right),
\end{equation}
where $q_n$ is the $n$-th query. 

\crd{
\section{Benchmark for open-vocabulary language-driven 3D retrieval}
\label{sec:retrieval}
In~\autoref{table:retrieval_counts}, we present queries contained in the retrieval benchmark and the number of queries in individual splits.
}
\begin{table}[t]
\scriptsize
    \centering
    \rowcolors{2}{gray!10}{white}
    \caption{
    \crd{
    \textbf{Number of queries in the individual splits of the retrieval benchmark dataset}.
    }
    }
    \begin{tabular}{l|cccc}
query & val & test & train\\
\hline\hline
agitator truck & 0 & 0 & 1 & 1\\
bulldozer & 0 & 1 & 1 & 2\\
excavator & 3 & 1 & 1 & 5\\
asphalt roller & 0 & 0 & 1 & 1\\
dustcart & 1 & 0 & 0 & 1\\
boom lift vehicle & 0 & 1 & 4 & 5\\
sedan & 1 & 0 & 0 & 1\\
sports car & 1 & 0 & 0 & 1\\
hatchback & 1 & 0 & 0 & 1\\
mini-van & 0 & 0 & 1 & 1\\
van & 0 & 0 & 1 & 1\\
lorry & 0 & 0 & 2 & 2\\
wagon & 0 & 0 & 1 & 1\\
SUV & 1 & 1 & 0 & 2\\
jeep & 1 & 1 & 0 & 2\\
campervan & 0 & 0 & 1 & 1\\
motorcycle & 0 & 0 & 1 & 1\\
vespa with driver & 0 & 0 & 1 & 1\\
golf cart & 0 & 1 & 0 & 1\\
forklift & 0 & 0 & 1 & 1\\
scooter with rider & 0 & 0 & 1 & 1\\
skateboard with rider & 0 & 1 & 0 & 1\\
ice cream van & 0 & 1 & 0 & 1\\
parcel delivery vehicle & 1 & 1 & 0 & 2\\
food truck & 0 & 0 & 1 & 1\\
police car & 0 & 1 & 0 & 1\\
police van & 0 & 0 & 1 & 1\\
dog & 0 & 0 & 1 & 1\\
bird & 0 & 0 & 2 & 2\\
double decker bus & 1 & 0 & 1 & 2\\
pick up truck for human transport & 0 & 0 & 1 & 1\\
jogger & 0 & 0 & 1 & 1\\
stroller & 0 & 0 & 1 & 1\\
two persons walking together & 1 & 0 & 0 & 1\\
person with a leaf blower & 0 & 0 & 1 & 1\\
chair & 0 & 1 & 0 & 1\\
stairs & 1 & 0 & 1 & 2\\
horse sculpture & 1 & 0 & 0 & 1\\
vase & 0 & 0 & 1 & 1\\
traffic lights & 0 & 0 & 1 & 1\\
fire hydrant & 1 & 2 & 1 & 4\\
mailbox & 3 & 0 & 0 & 3\\
mailboxes & 0 & 0 & 1 & 1\\
suitcase & 0 & 0 & 1 & 1\\
wheelbarrow & 0 & 0 & 1 & 1\\
garbage bin & 0 & 0 & 1 & 1\\
cardboard box & 0 & 0 & 1 & 1\\
mirror & 0 & 1 & 0 & 1\\
human with an umbrella & 1 & 0 & 2 & 3\\
rain barrel & 0 & 0 & 1 & 1\\
mobile toilet & 1 & 0 & 0 & 1\\
pedestrian crossing & 1 & 0 & 0 & 1\\
barrier gate & 0 & 0 & 2 & 2\\
motorbike & 0 & 1 & 0 & 1\\
yellow school bus & 0 & 1 & 0 & 1\\
police officer & 0 & 2 & 0 & 2\\
chopper & 0 & 1 & 0 & 1\\
small bulldozer & 0 & 1 & 0 & 1\\
concrete mixer truck & 0 & 2 & 0 & 2\\
truck crane & 0 & 1 & 0 & 1\\
cabriolet & 0 & 1 & 0 & 1\\
yellow car & 0 & 1 & 0 & 1\\
baby stroller & 0 & 2 & 0 & 2\\
green trash bin & 0 & 1 & 0 & 1\\
red sedan & 1 & 1 & 0 & 2\\
delivery van & 0 & 1 & 0 & 1\\
white truck tractor & 0 & 1 & 0 & 1\\
keg barrels & 0 & 1 & 0 & 1\\
regular-cab truck & 0 & 1 & 0 & 1\\
minibus & 0 & 1 & 0 & 1\\
trash bin & 1 & 1 & 0 & 2\\
white suv & 1 & 0 & 0 & 1\\
delivery truck & 1 & 0 & 0 & 1\\
black truck with a trailer & 1 & 0 & 0 & 1\\
person on a bicycle & 1 & 0 & 0 & 1\\
scooter & 1 & 0 & 0 & 1\\
\hline
total & 28 & 35 & 42 & 105 \\
    \end{tabular}
    \label{table:retrieval_counts}
\end{table}
\section{Qualitative results}
\label{sec:supp_qual}

In this section, we first show additional qualitative results for the task of zero-shot \task ~using 16 classes in the nuScenes~\cite{caesar2020nuscenes} dataset
in~Figures~\ref{fig:supp_zeroshot0},~\ref{fig:supp_zeroshot1}~and~\ref{fig:supp_zeroshot2}. 
We further proceed with qualitative examples of the retrieval task 
in Figures~\ref{fig:supp_retrieval0}~and~\ref{fig:supp_retrieval1}.

\textbf{Zero-shot \task.}
\label{sec:supp_qual_zeroshot}
In Figures~\ref{fig:supp_zeroshot0},~\ref{fig:supp_zeroshot1},~and~\ref{fig:supp_zeroshot2}, we present qualitative results of zero-shot \task ~for 16 semantic categories in the nuScenes dataset~\cite{caesar2020nuscenes}. These figures showcase the ability of our method to reconstruct the overall 3D structure of the scene accurately. Moreover, as shown in Figure~\ref{fig:supp_zeroshot1}, our method can recognize classes such as \textit{bus}, which are not well represented in the training dataset.

\begin{figure}[t]
    \centering
    \includegraphics[width=1.0\linewidth]{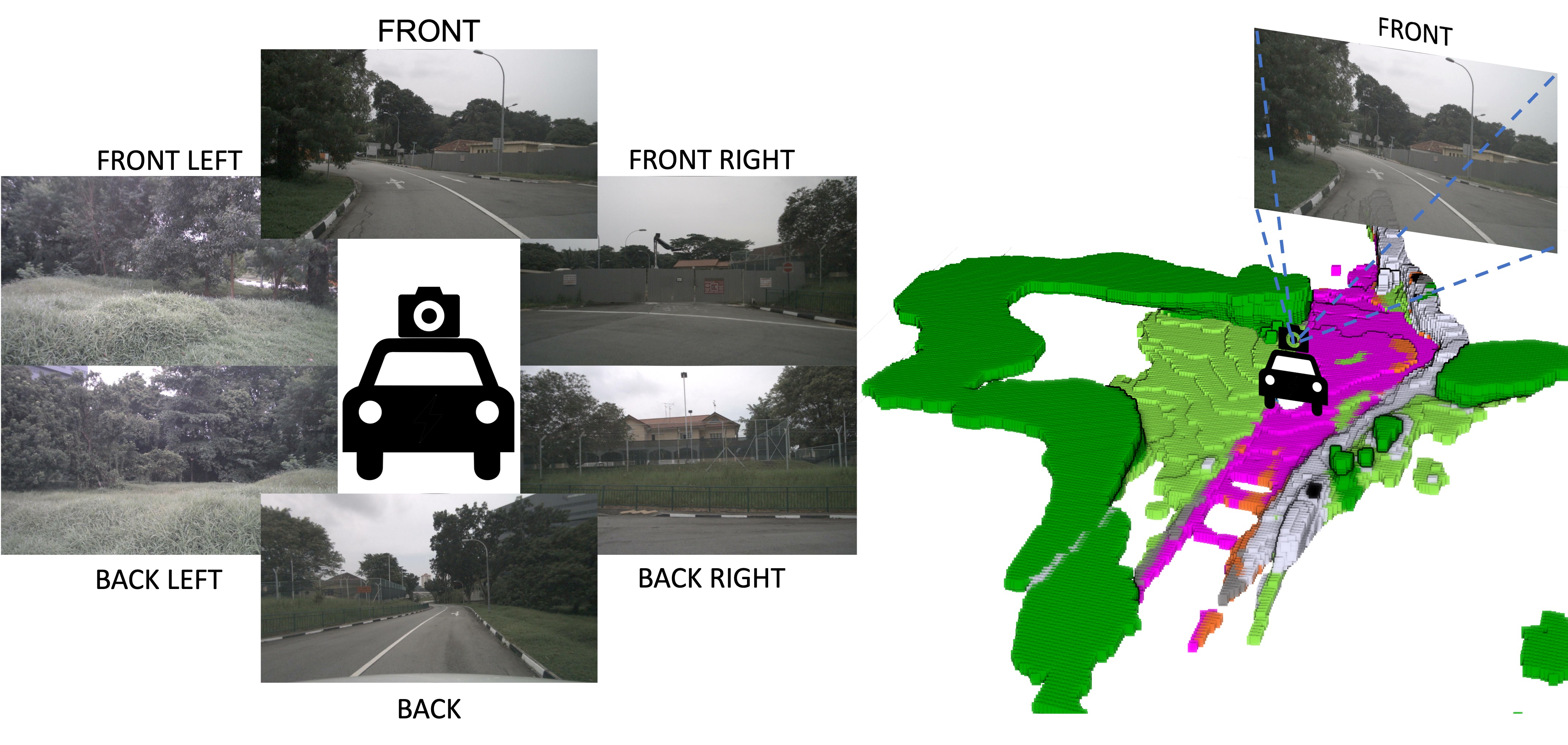}\\
    \includegraphics[width=1.0\linewidth]{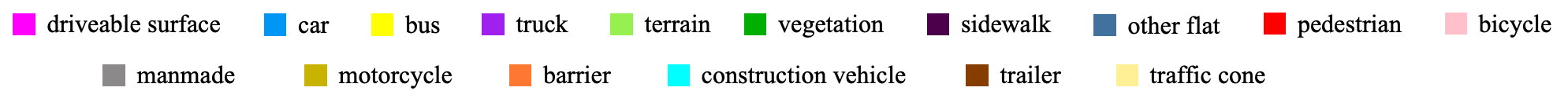}\\
    \caption{
    \textbf{Zero-shot \task}. \textit{Left:} six input surround-view images. 
    \textit{Right:} our prediction; training grid resolution 100$\times$100$\times$8 is upsampled to 300$\times$300$\times$24 by interpolating the trained representation space.
    }
    \label{fig:supp_zeroshot0}
\end{figure}

\begin{figure}[t]
    \centering
    \includegraphics[width=1.0\linewidth]{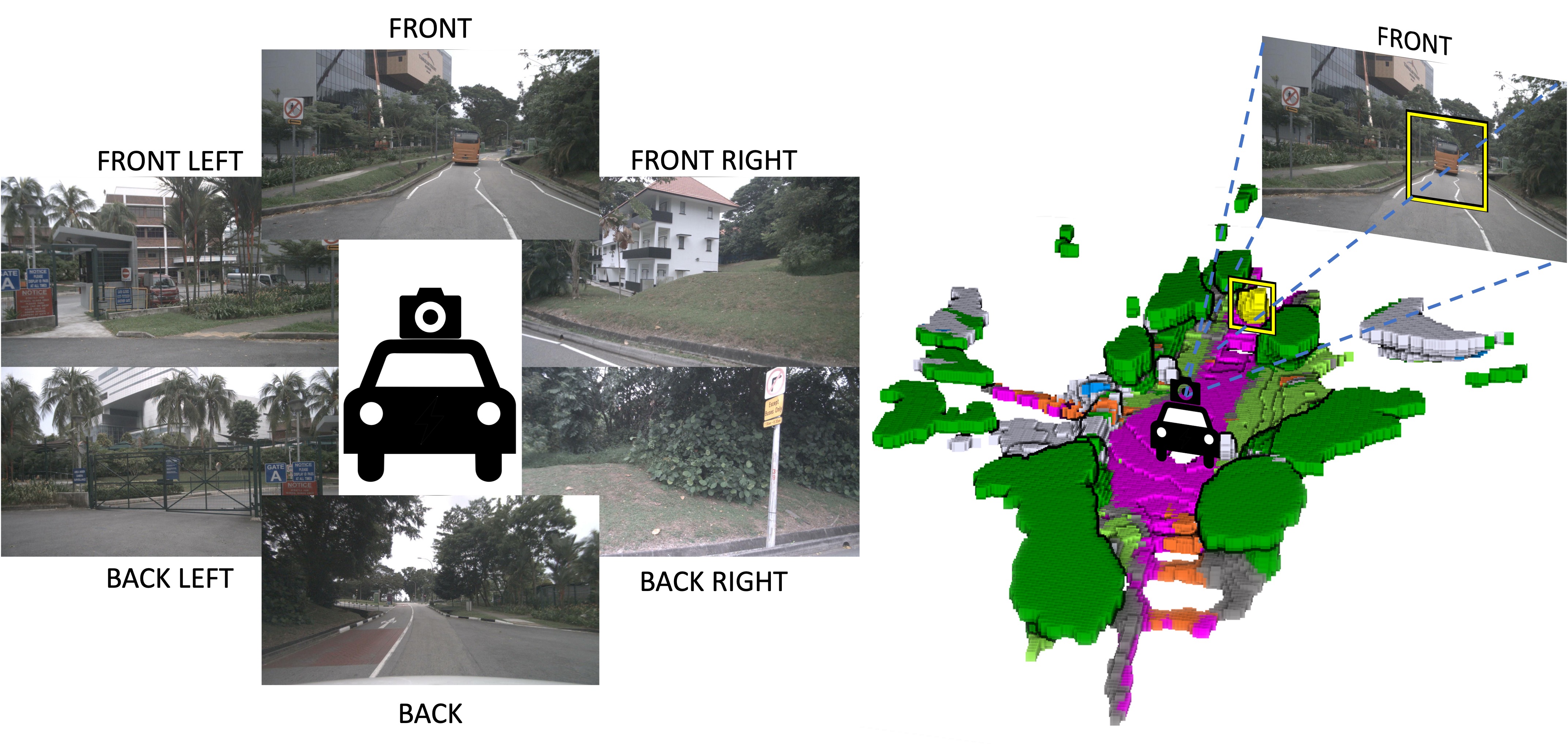}\\
    \includegraphics[width=1.0\linewidth]{images/legend_ours.png}\\
        \caption{
    \textbf{Zero-shot \task}. \textit{Left:} six input surround-view images. 
    \textit{Right:} our prediction; training grid resolution 100$\times$100$\times$8 is upsampled to 300$\times$300$\times$24 by interpolating the trained representation space. It is worth noting that the model successfully segments even the class \textit{bus}, despite its limited occurrence in the training set.
    }
    \label{fig:supp_zeroshot1}
\end{figure}

\begin{figure}
    \centering
    \includegraphics[width=0.33\linewidth]{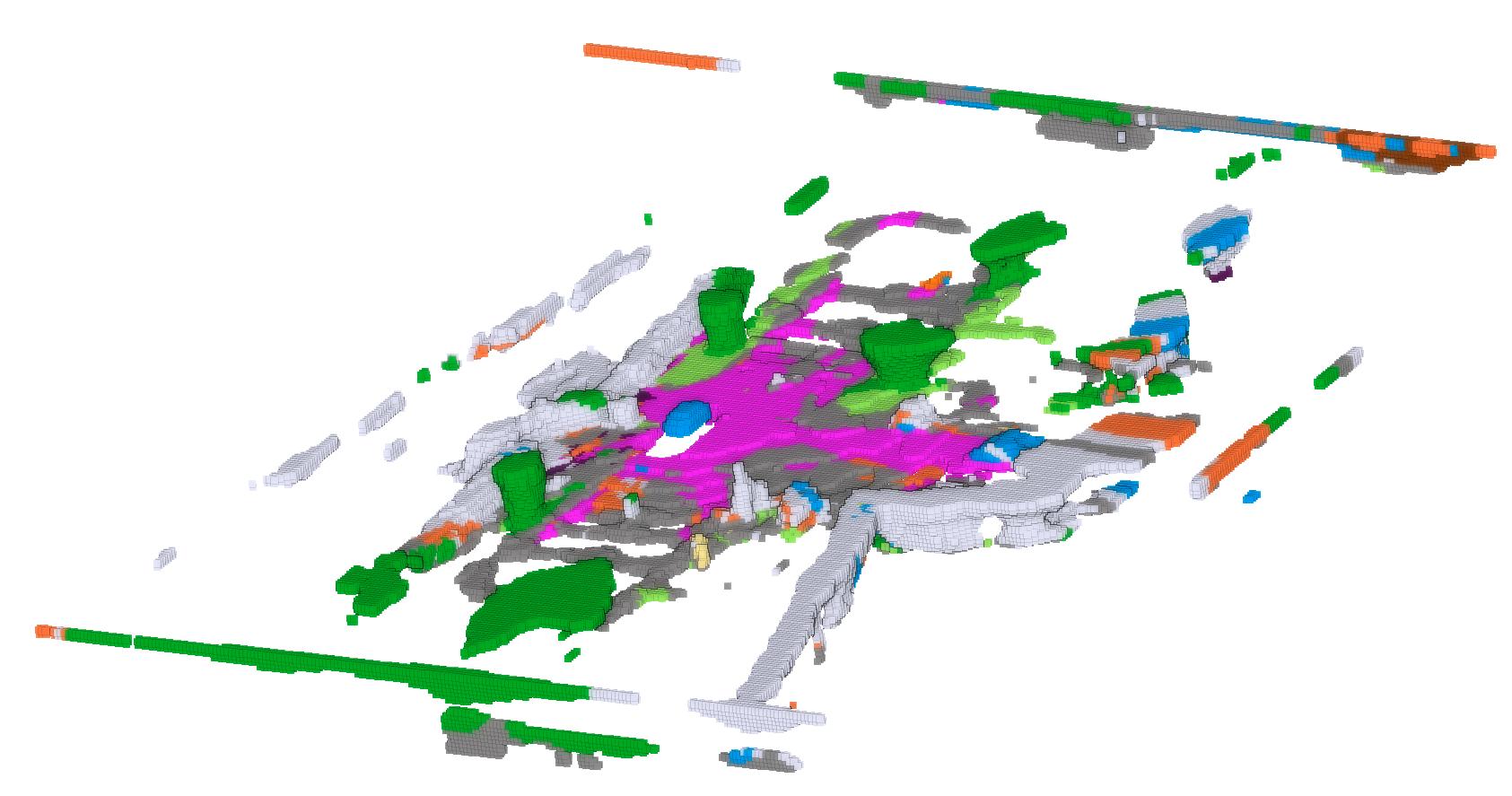}%
    \includegraphics[width=0.33\linewidth]{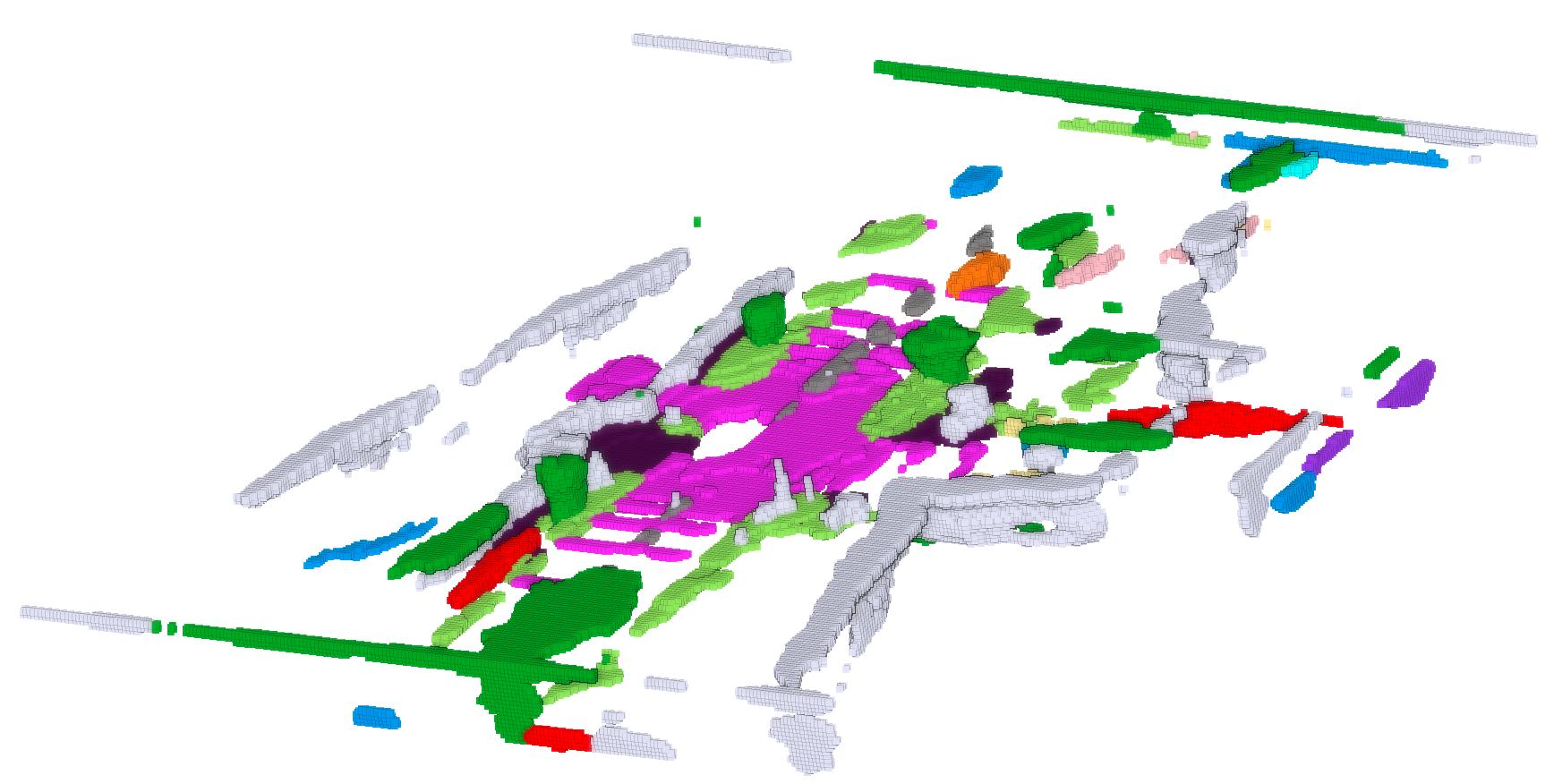}%
    \includegraphics[width=0.33\linewidth]{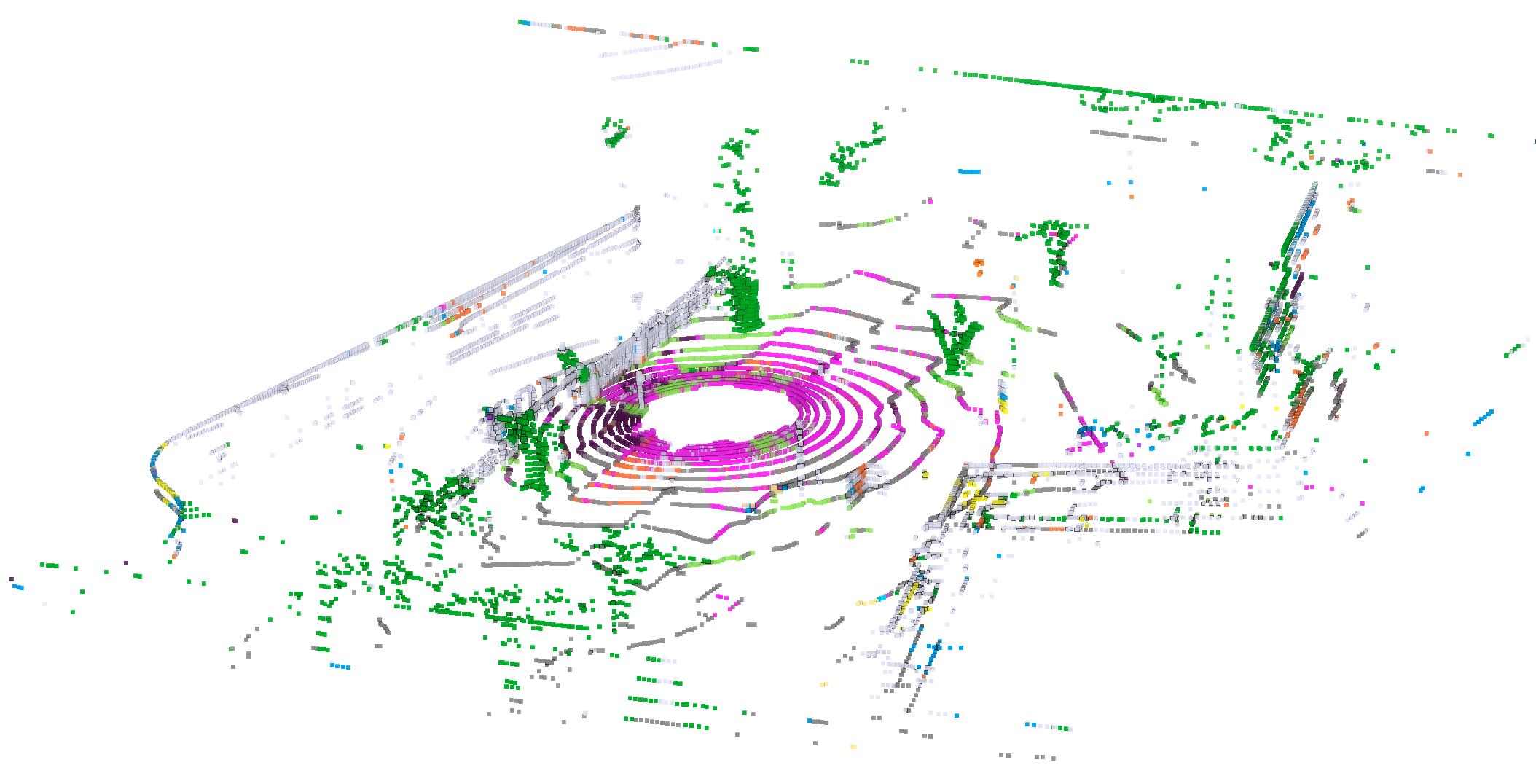}\\
    \includegraphics[width=0.33\linewidth]{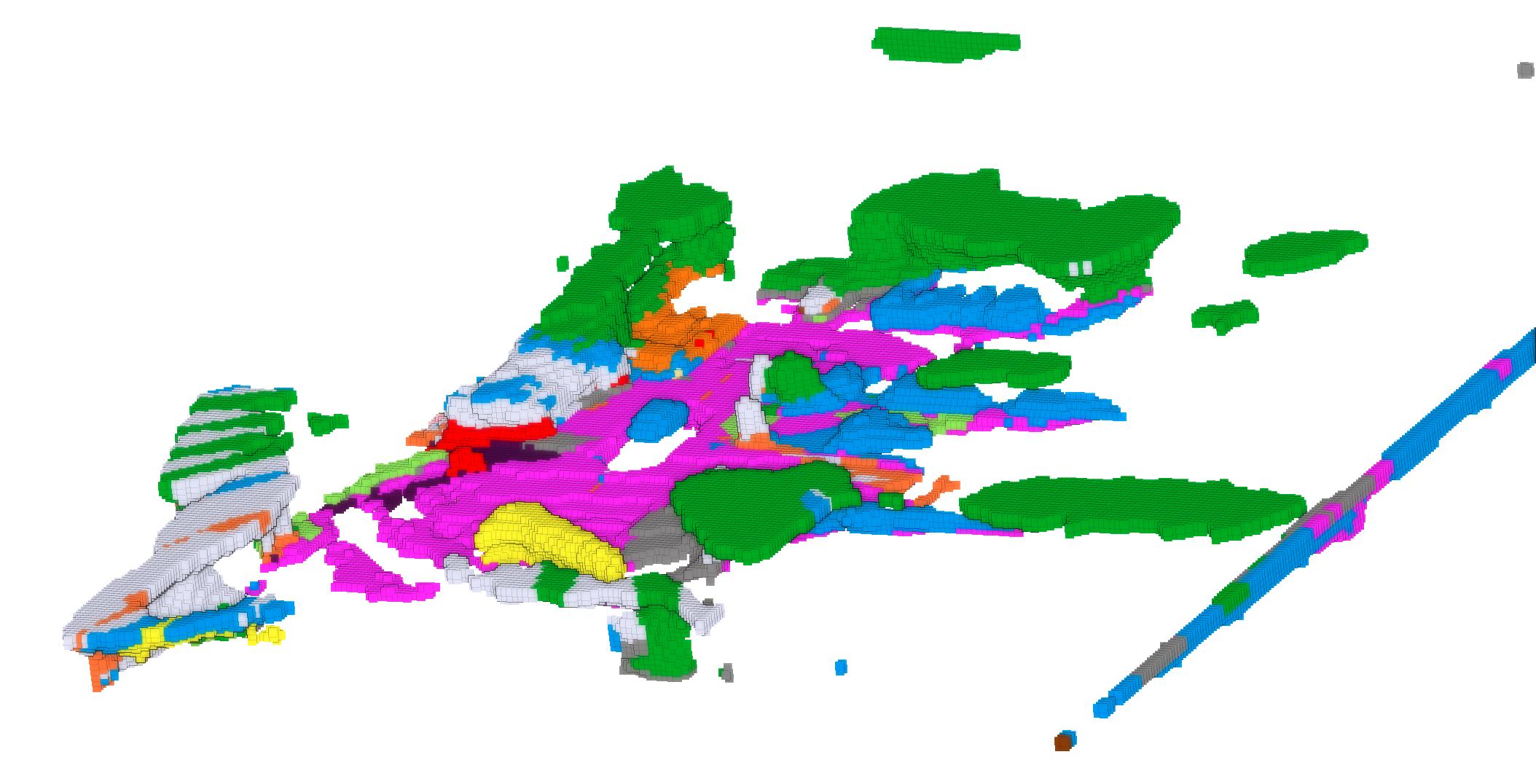}%
    \includegraphics[width=0.33\linewidth]{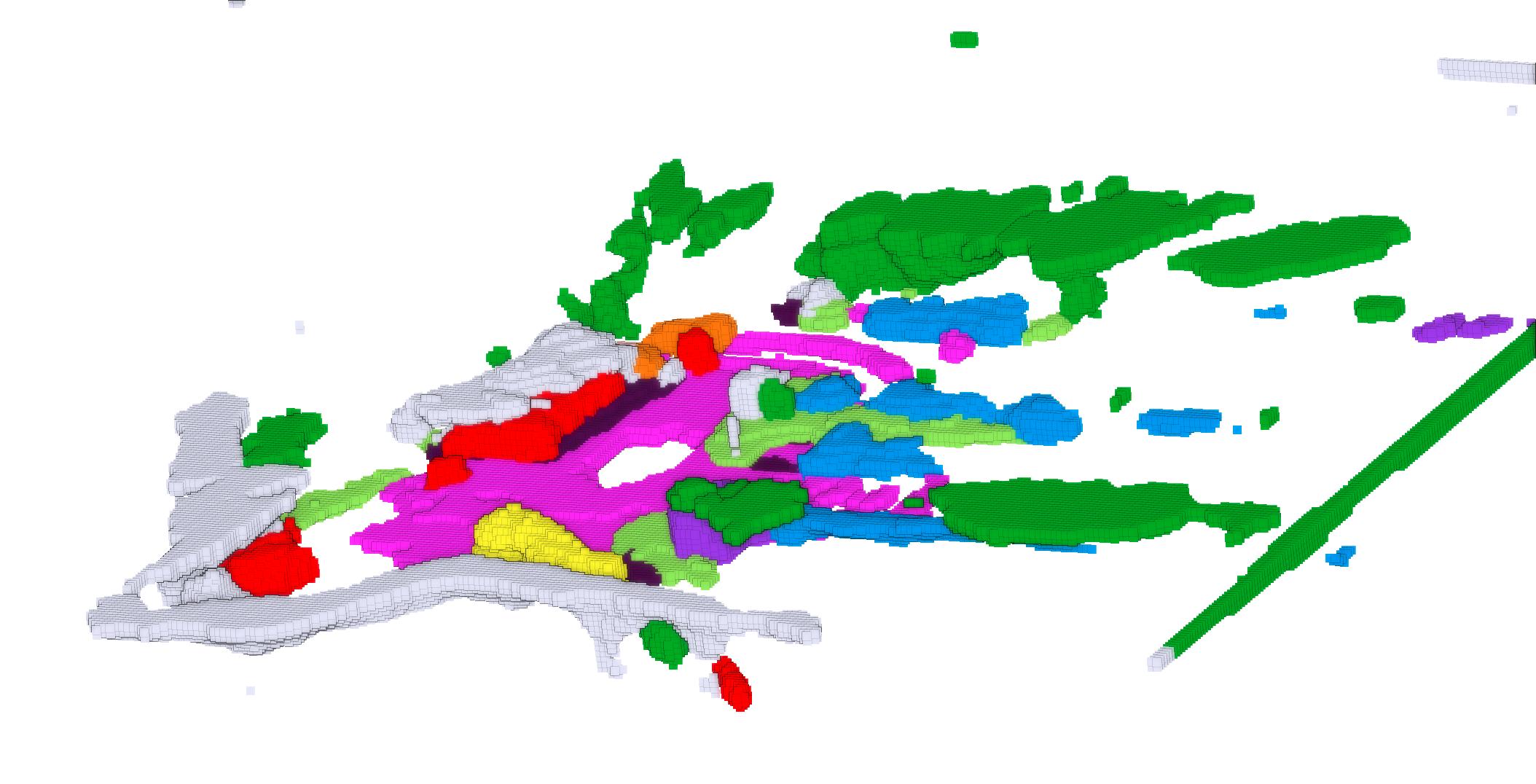}%
    \includegraphics[width=0.33\linewidth]{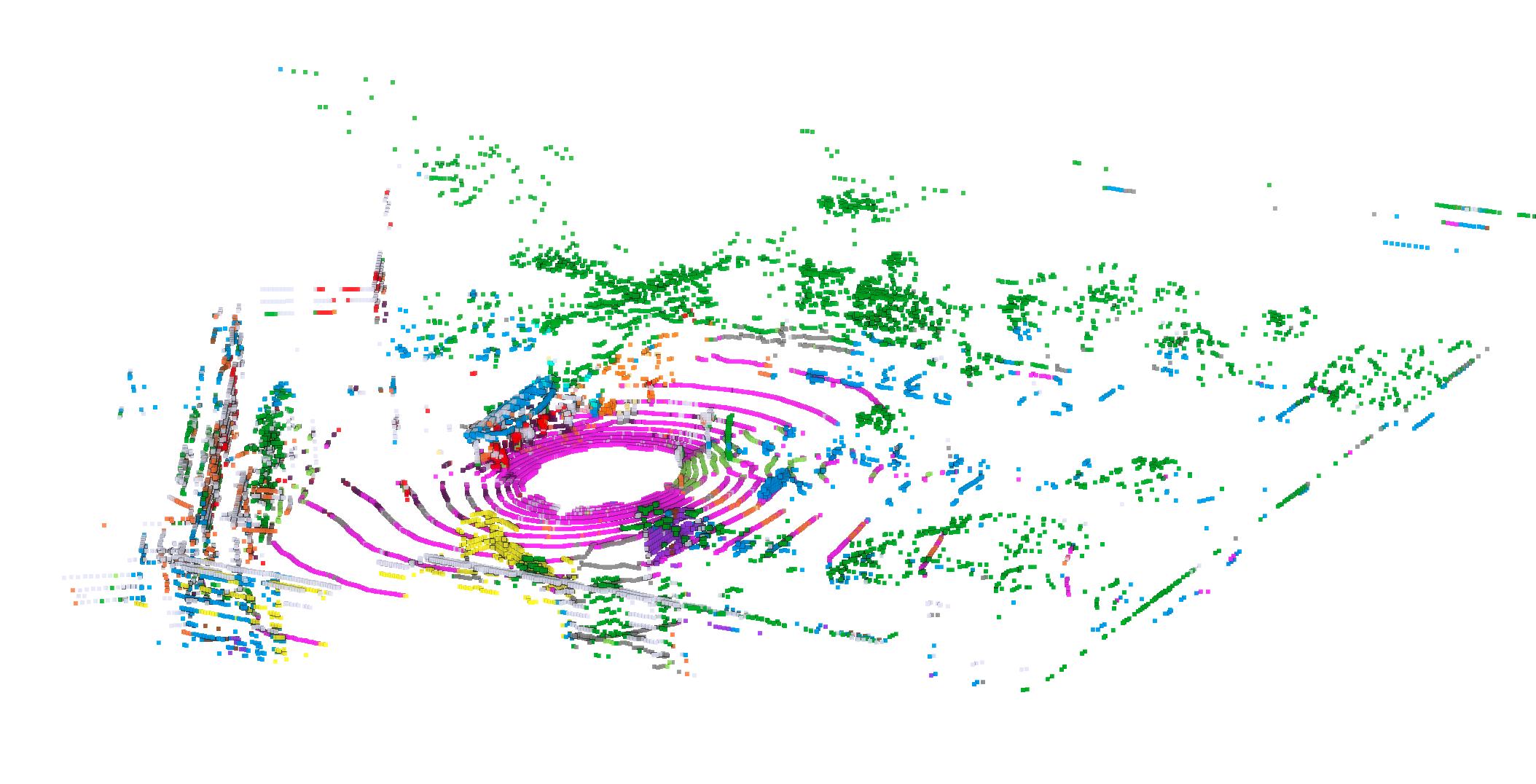}\\
    \includegraphics[width=0.33\linewidth]{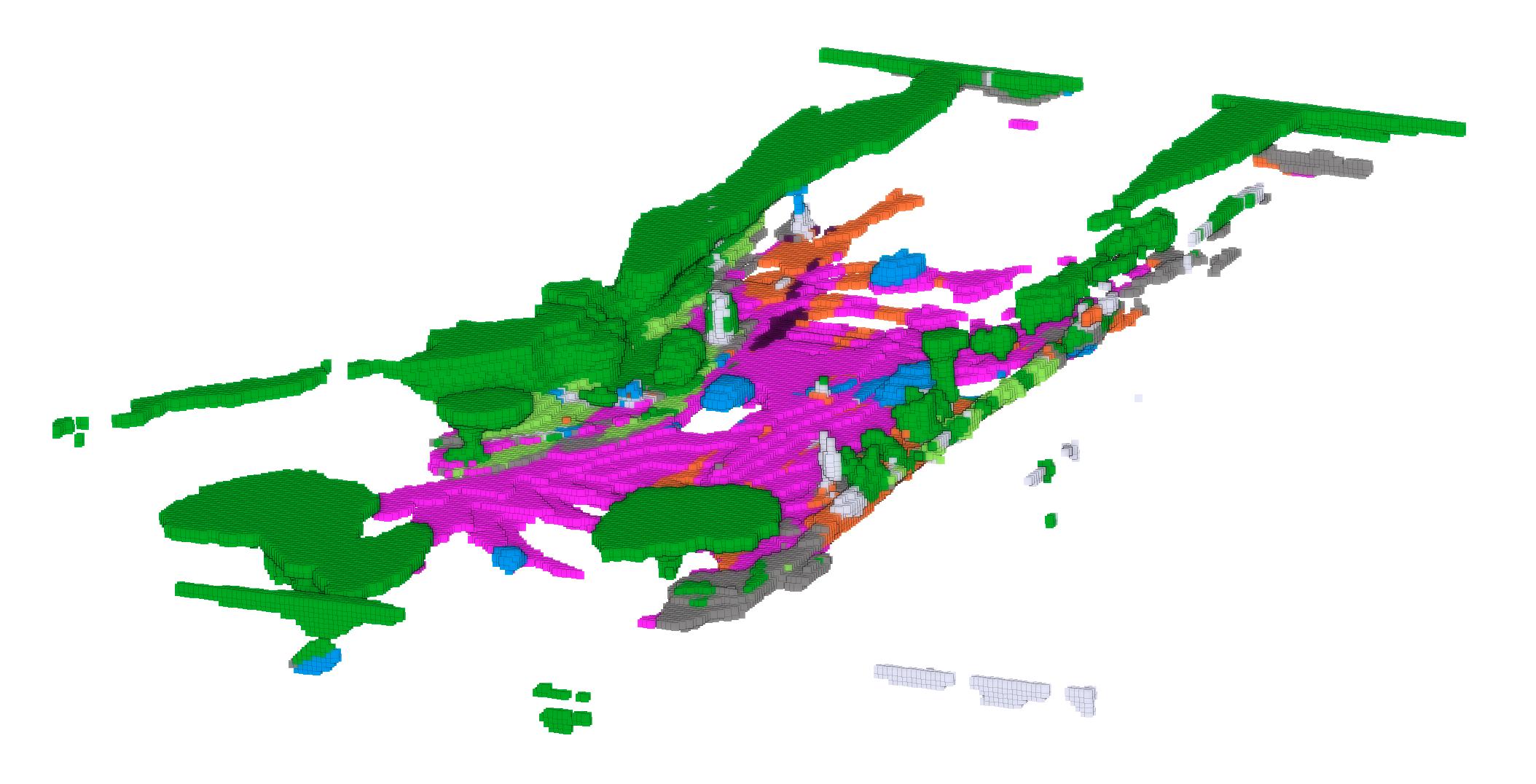}%
    \includegraphics[width=0.33\linewidth]{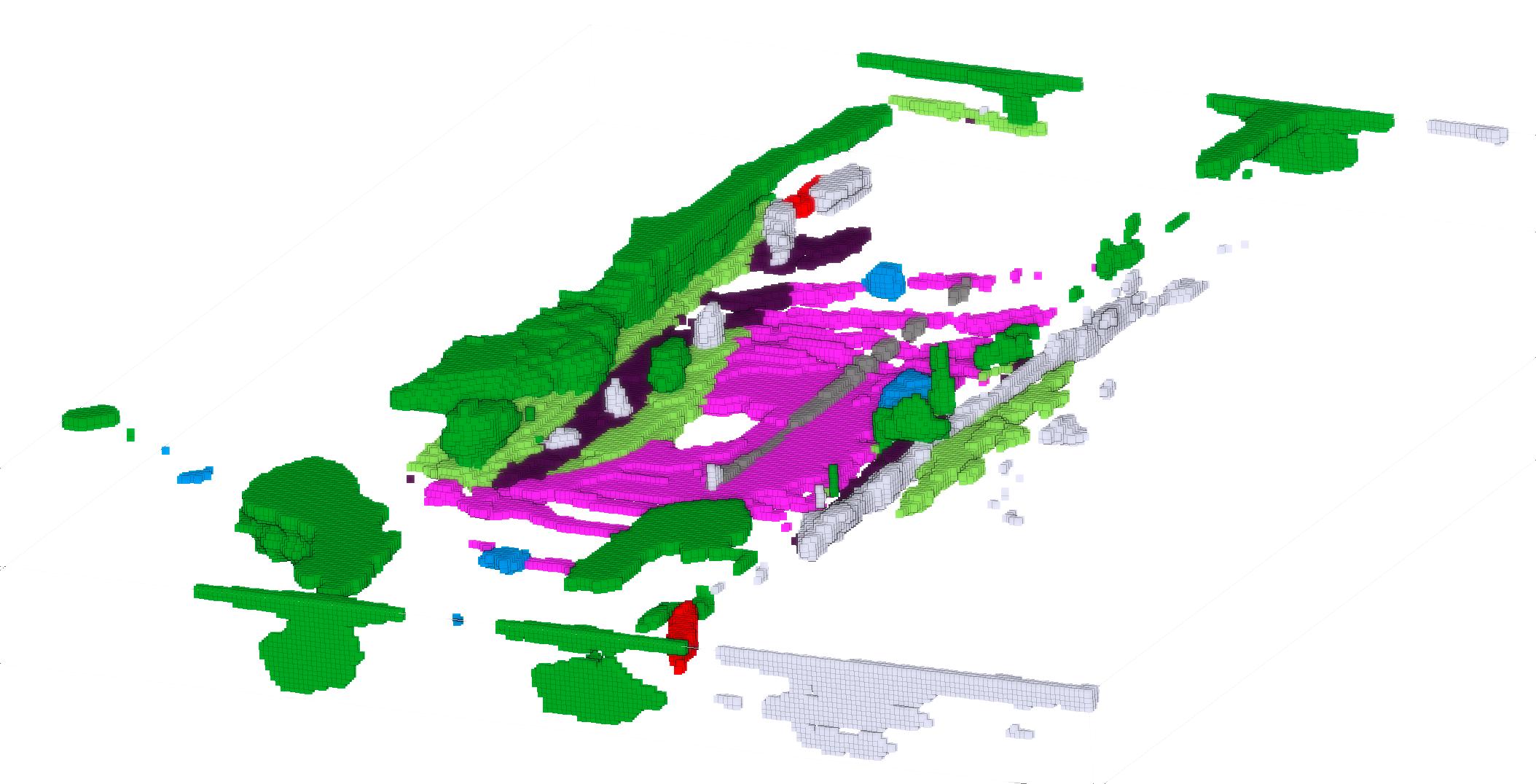}%
    \includegraphics[width=0.33\linewidth]{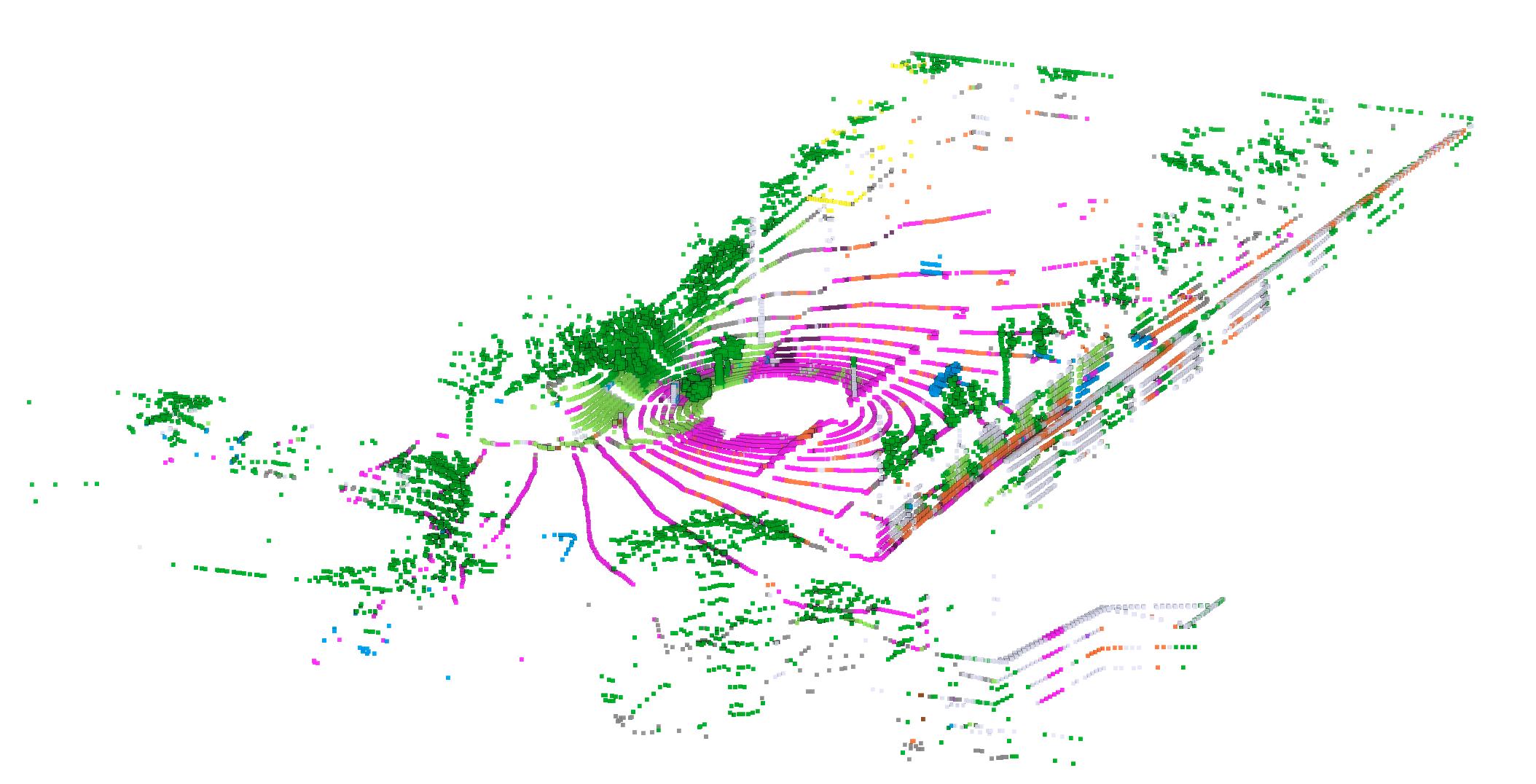}\\
     \begin{tabular}{ccc}
         \textbf{\hspace{5ex} \ours~(ours)} \hspace{15ex}  & \hspace{2ex} {TPVFormer}  \hspace{10ex} & {MaskCLIP+} \\
         && {projected to GT point cloud}
    \end{tabular}
    \caption{
    \crd{
    \textbf{Qualitative results.} Qualitative comparison of our \ours~to fully-supervised TPVFormer and to MaskCLIP+ results projected from 2D to 3D ground-truth point cloud.
    }
    }
    \label{fig:qual_supp_2}
\end{figure}

\begin{figure}[t]
    \centering
    \includegraphics[width=1.0\linewidth]{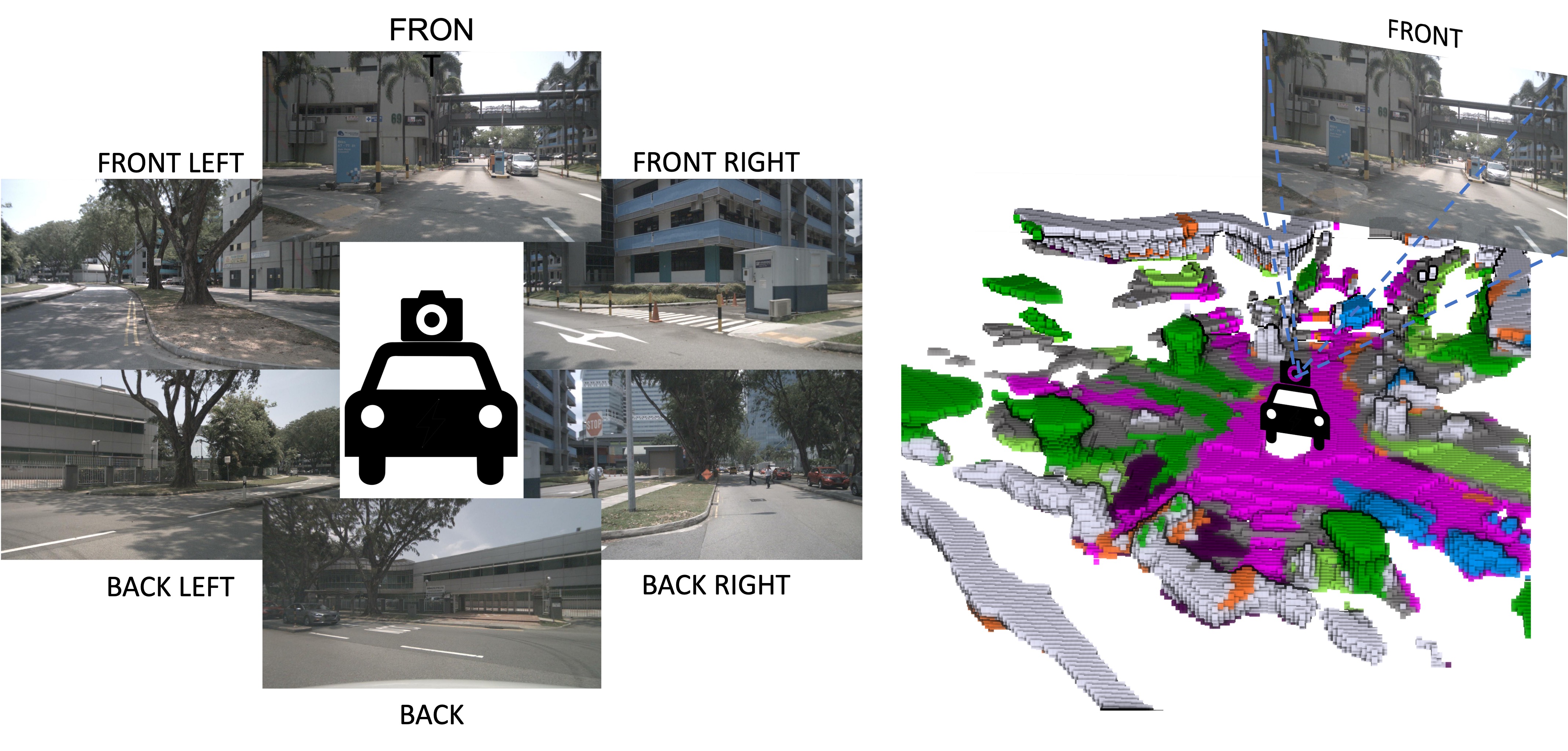}\\
    \includegraphics[width=1.0\linewidth]{images/legend_ours.png}\\
    
        \caption{
    \textbf{Zero-shot \task}. \textit{Left:} six input surround-view images. 
    \textit{Right:} our prediction; training grid resolution 100$\times$100$\times$8 is upsampled to 300$\times$300$\times$24 by interpolating the trained representation space. This example demonstrates the model's ability to accurately reconstruct complex 3D scenes.
    }
    \label{fig:supp_zeroshot2}
\end{figure}

\textbf{Text-based retrieval.}
\label{sec:supp_qual_retrieval}
We present qualitative results of the retrieval task in Figures~\ref{fig:supp_retrieval0} and \ref{fig:supp_retrieval1}. These figures demonstrate the effectiveness of our model in retrieving non-annotated categories, such as \textit{stairs} or \textit{zebra crossing}, by querying the predicted features with a single text query and visualizing the similarity overlaid on the voxel grid. However, we observed limitations in the retrieval capabilities due to two factors: a) the resolution of the voxel grid and b) the level of concept granularity captured in MaskCLIP+ features. 
It is important to note that although MaskCLIP+ features (extracted with a DeepLabv2 architecture) have better spatial precision, they do not fully preserve all the descriptive capabilities of the original CLIP, as pointed out also in~\cite{conceptfusion} since they are already distilled from the CLIP model.

\crd{
\textbf{Qualitative comparison.} In~\autoref{fig:qual_supp_2}, we present a qualitative comparison of our \ours~to fully-supervised TPVFormer and to MaskCLIP+ results projected from 2D to 3D ground-truth point cloud.
}

\begin{figure}
    \centering
    \includegraphics[width=0.32\linewidth]{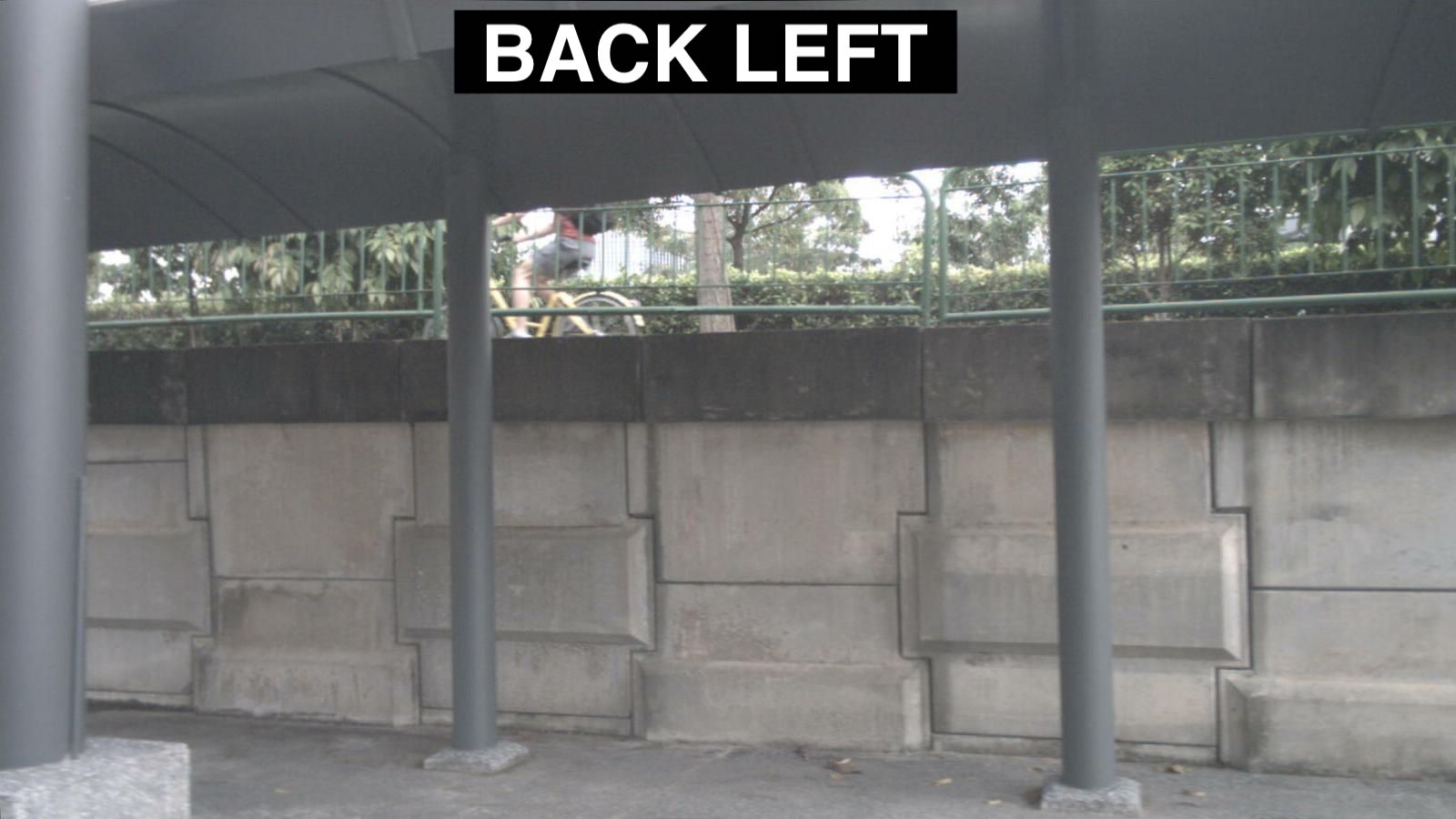}
    \includegraphics[width=0.32\linewidth]{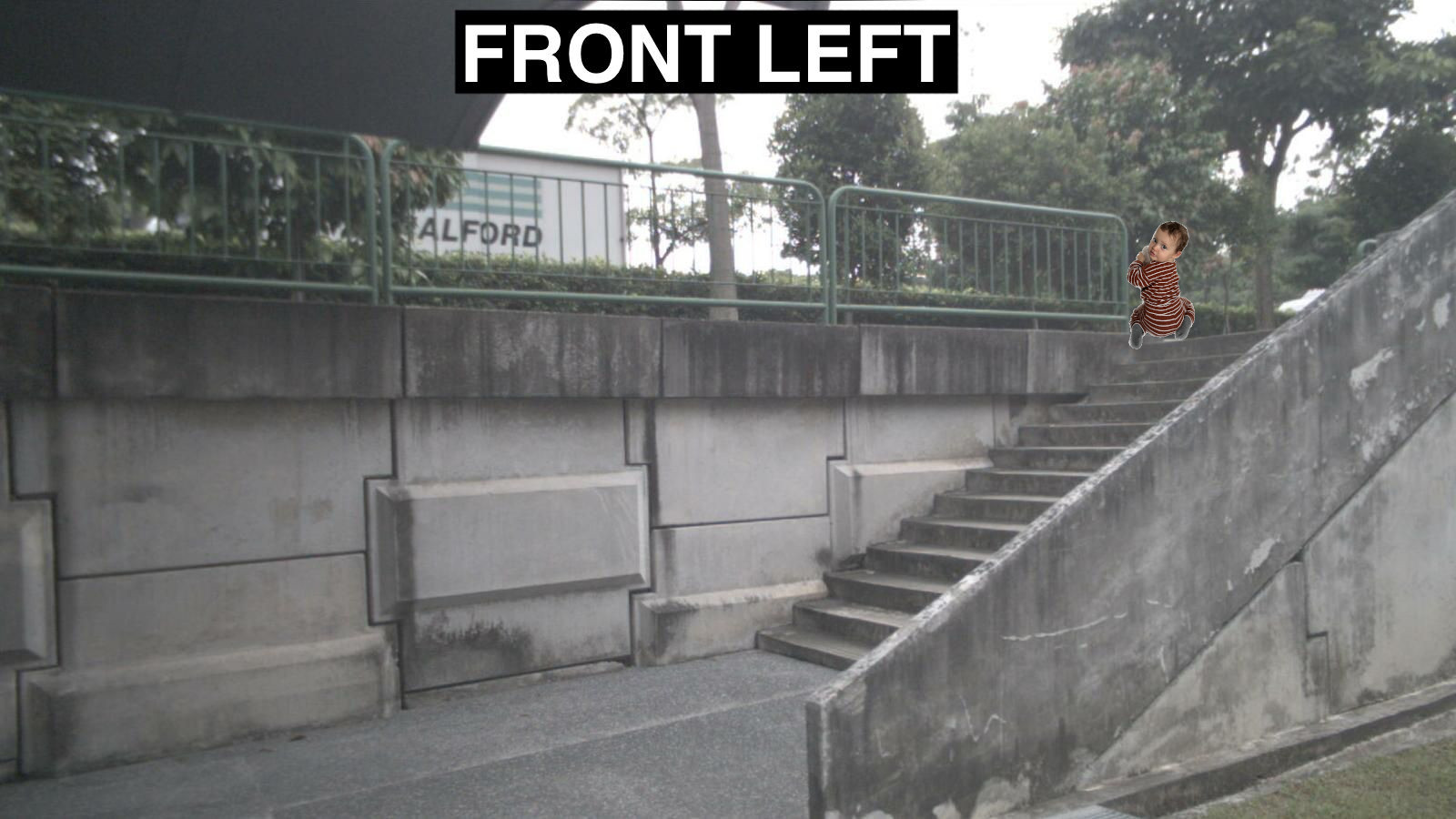}
    \includegraphics[width=0.32\linewidth]{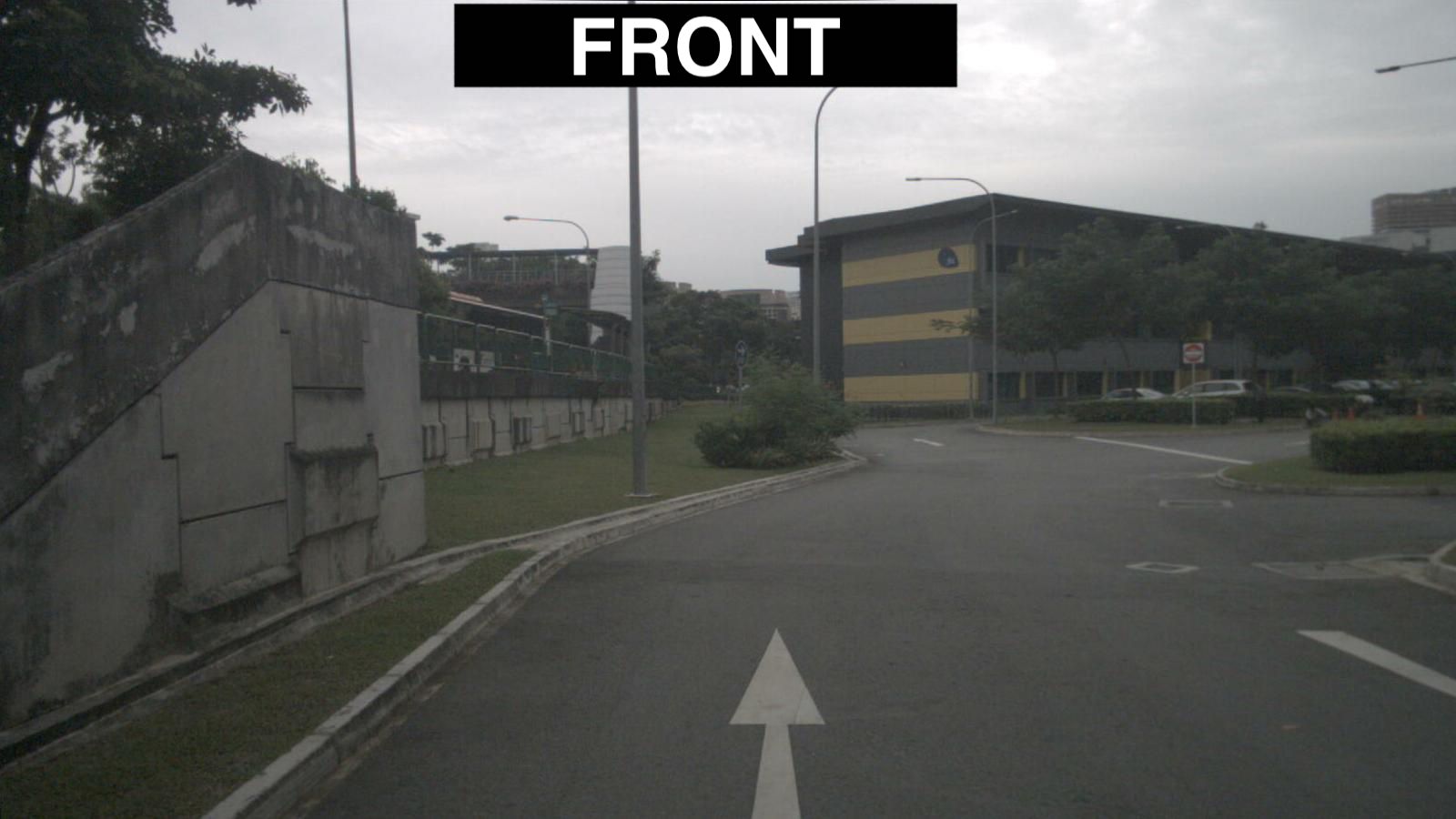}\\
    \includegraphics[width=1.0\linewidth]{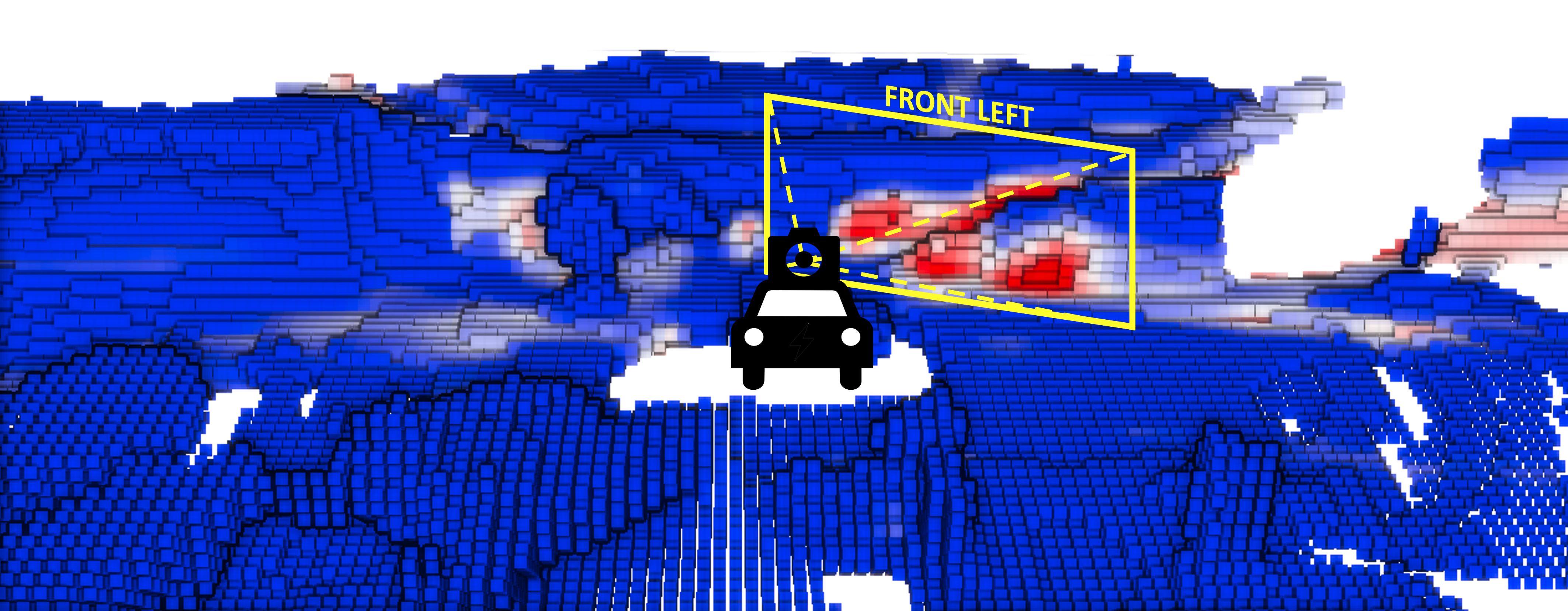}
    \caption{\textbf{Text-based 3D retrieval.} Query: `\textit{stairs}'. Top: Input images showing just the left side of the car. Bottom: a detailed view of the corresponding 3D scene with a heatmap indicating the similarity to the `\textit{stairs}' query. The stairs are mostly visible in the front left camera, as also highlighted in the predicted 3D~scene.}
    \label{fig:supp_retrieval0}
\end{figure}

\begin{figure}
    \centering
    \includegraphics[width=0.495\linewidth]{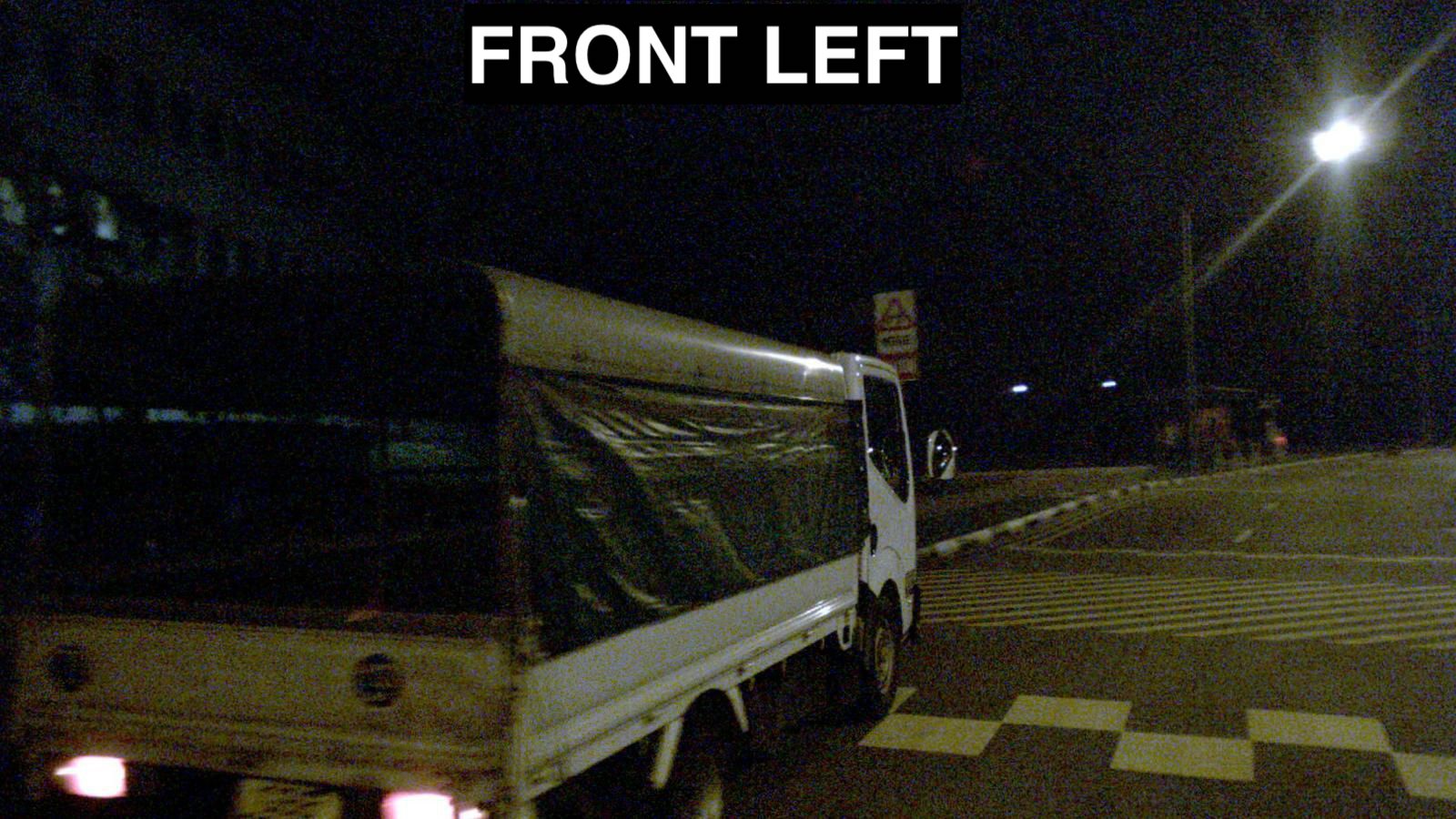}
    \includegraphics[width=0.495\linewidth]{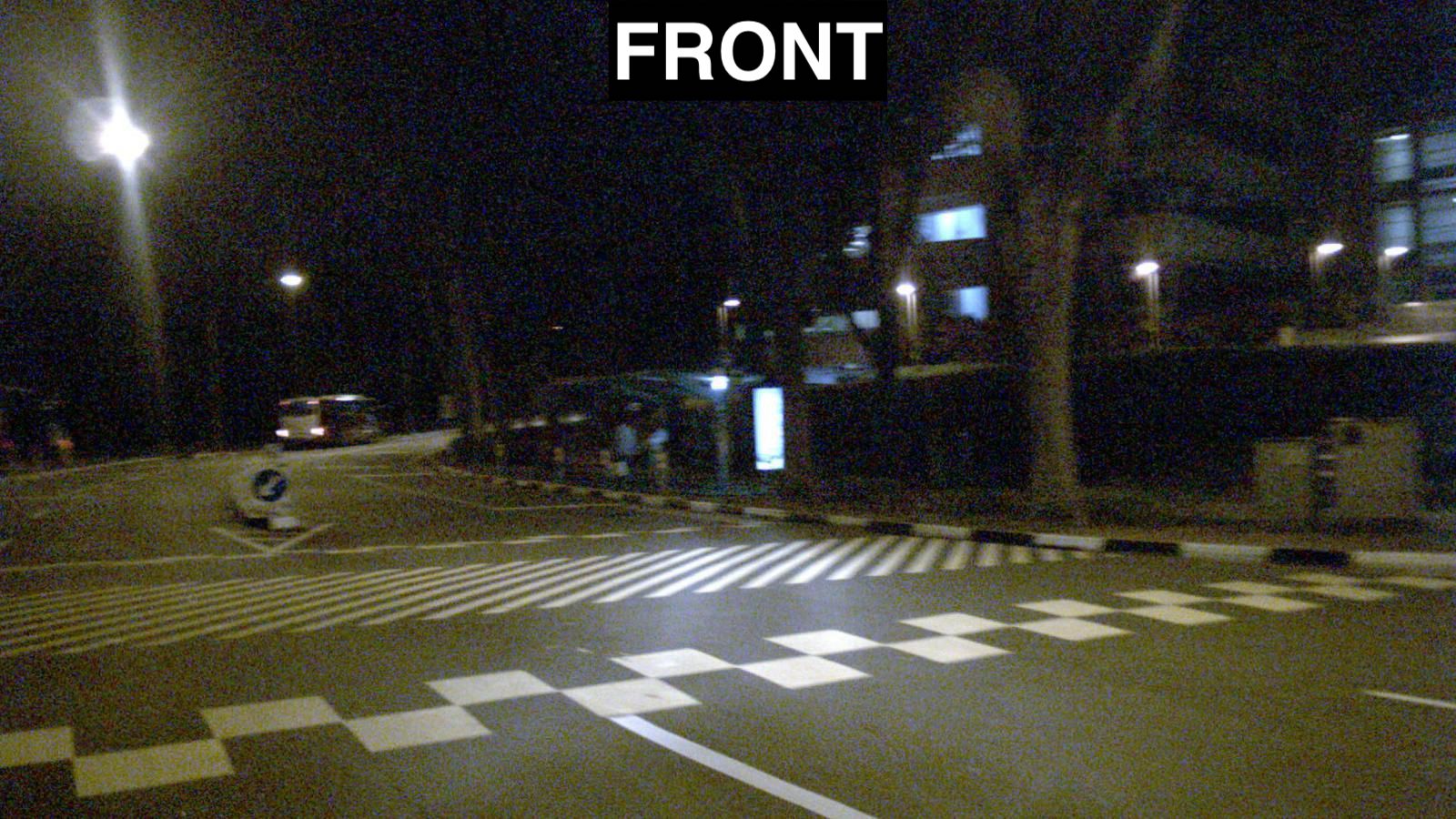}\\
    \includegraphics[width=0.995\linewidth]{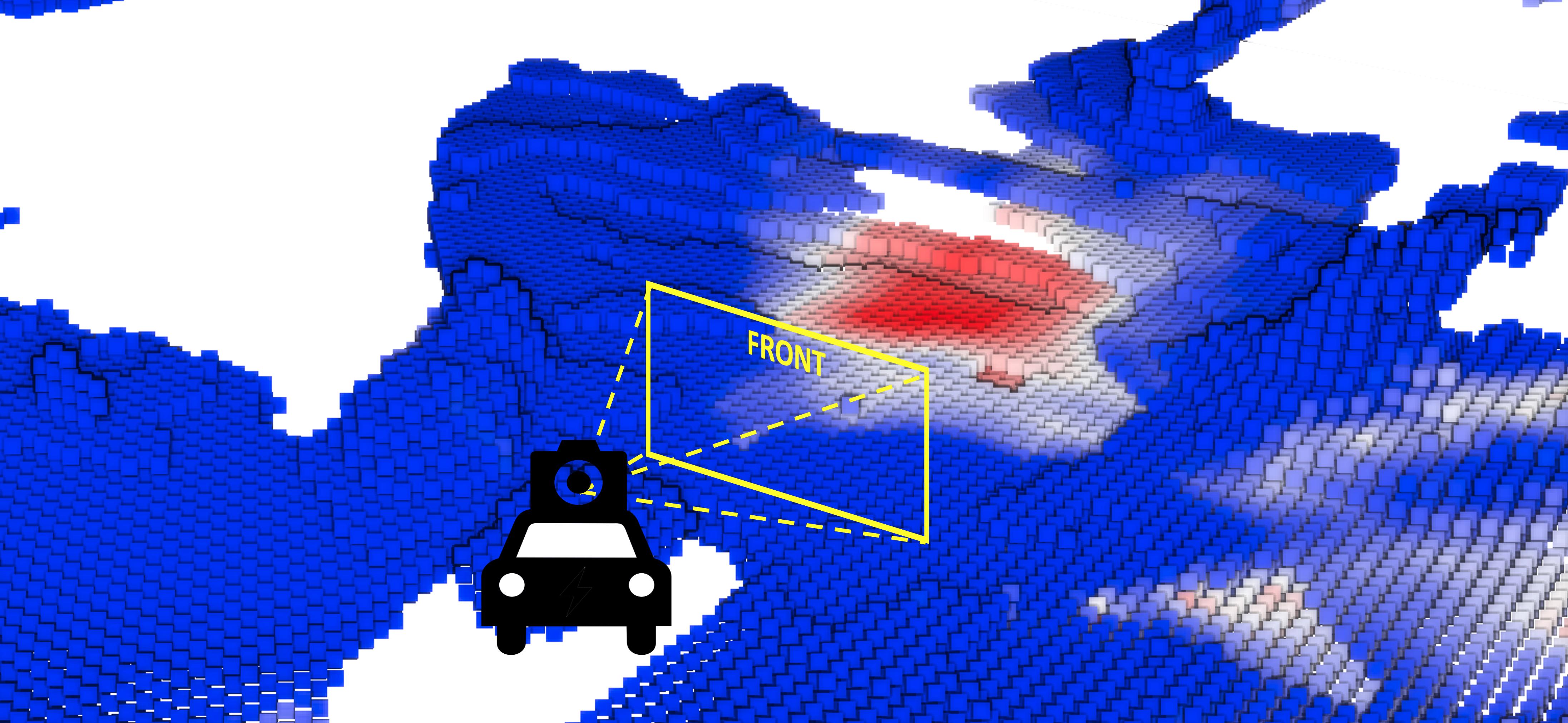}
    \caption{\textbf{Text-based 3D retrieval.} Query: `\textit{zebra crossing}'. Top: Input images showing the scene in front of the car. Bottom: a detailed view of the corresponding 3D scene with a heatmap indicating the similarity to the `\textit{zebra crossing}' query. This example showcases the model's ability to recognize fine-grained concepts, even under challenging conditions like nighttime.}
    \label{fig:supp_retrieval1}
\end{figure}

\end{document}